\documentclass[twoside]{article}
\usepackage{ecj,palatino,epsfig,latexsym, amsmath, algorithm, algpseudocodex, array, adjustbox, textcomp, stfloats, url, verbatim, graphicx, bm, makecell, amsfonts, amssymb, caption}
\usepackage[authoryear, round]{natbib}  
\usepackage{longtable}
\usepackage{pdflscape} 
\usepackage{array} 
\usepackage{booktabs}
\usepackage{xcolor}

\definecolor{changeColor}{rgb}{0, 0, 0} 
\definecolor{changeColor2}{rgb}{0, 0, 0}
\newcommand{\changea}[1]{\textcolor{changeColor}{#1}}
\newcommand{\changeb}[1]{\textcolor{changeColor2}{#1}}

\begin{document}

\ecjHeader{x}{x}{xxx-xxx}{202X}{Improving CMA-ES in the presence of noise}{R. M. Martin and S. H. Collins}
\title{\bf Improving CMA-ES convergence speed, efficiency, and reliability in noisy robot optimization problems}  

\author{\name{\bf Russell M. Martin} \hfill \addr{rumartin@stanford.edu}\\
\addr{Department of Mechanical
Engineering, Stanford University, Stanford, 94305, USA}
\AND
\name{\bf Steven H. Collins} \hfill \addr{stevecollins@stanford.edu}\\
\addr{Department of Mechanical
Engineering, Stanford University, Stanford, 94305, USA}
}

\maketitle

\begin{abstract}

Experimental robot optimization often requires evaluating each candidate policy for seconds to minutes. The chosen evaluation time influences optimization because of a speed-accuracy tradeoff: shorter evaluations enable faster iteration, but are also more subject to noise. Here, we introduce a supplement to the CMA-ES optimization algorithm, named Adaptive Sampling CMA-ES (AS-CMA), which assigns sampling time to candidates based on predicted sorting difficulty, aiming to achieve consistent precision. We compared AS-CMA to CMA-ES and Bayesian optimization using a range of static sampling times in four simulated cost landscapes. AS-CMA converged on 98\% of all runs without adjustment to its tunable parameter, and converged 24-65\% faster and with 29-76\% lower total cost than each landscape's best CMA-ES static sampling time. As compared to Bayesian optimization, AS-CMA converged more efficiently and reliably in complex landscapes, while in simpler landscapes, AS-CMA was less efficient but equally reliable. We deployed AS-CMA in an exoskeleton optimization experiment and found the optimizer’s behavior was consistent with expectations. These results indicate that AS-CMA can improve optimization efficiency in the presence of noise while minimally affecting optimization setup complexity and tuning requirements.

\end{abstract}


\section{Introduction}

For robotic systems that operate in complex and variable contexts, such as autonomous vehicles, dexterous manipulators, and collaborative robots, exclusively optimizing design parameters and control strategies in simulated environments can be insufficient because these simulations often do not fully capture the intricacies of the real world \citep{kang2019generalization,zhu2020ingredients, zhang2019vr, luo2024precise, peng2025data}. Alternatively, experimental optimization can be conducted directly with the physical robotic systems in their real environments. This confers the advantage of removing the need for the experimenter to parameterize details about the simulated environment, which can save time and, by testing in the real world instead of using a model, can lead to higher-quality optimization results. Experimental optimization techniques are also advantageous because they can enable continuous online improvement when integrated into a robot that has already been deployed \citep{zhu2020ingredients, slade_personalizing_2022}. 

One example domain of where real-world robot optimization has been advantageous is in optimization of exoskeletons that assist walking. Here, simulated optimization is inadequate because the human-robot interaction is difficult to model accurately: human response to robotic assistance is very difficult to predict  \citep{handford_energy-optimal_2018}, people can respond differently to the same assistance \citep{grimmer_comparison_2019}, and users' proficiency varies with time \citep{poggensee_how_2021}. Strategies for experimentally optimizing exoskeleton assistance range from hand-tuning, which delivers moderate benefits \citep{ding_effect_2016, malcolm_simple_2013}, to the deployment of algorithmic strategies known as human-in-the-loop optimization, which have produced significantly greater benefits in recent years \citep{ding_effect_2016, slade_personalizing_2022, zhang2017human}. Human-in-the-loop optimization typically involves the selection of exoskeleton timing and torque parameters while using human biomechanical measurements such as energy cost, muscle activity, or kinematics as the objective function. Although human-in-the-loop optimization has primarily been deployed in biomechatronic research, it is expected to have similar applicability to the optimization of industrial, medical, and transportation-related human-robot systems \citep{slade_human---loop_2024}. 

A challenge when conducting experimental optimization is the presence of noise in the signal(s) of interest, which makes optimization time-consuming and less precise \citep{arnold_general_2006, arnold_local_2001}. A simple yet useful model of noise is as follows: for a given candidate $x_i$, the influence of noise $\epsilon$ on measured cost $f_{meas}(x_i)$ relative to the true cost $f_{true}(x_i)$ is $f_{meas}(x_i) = f_{true}(x_i) + \epsilon$, where $\epsilon \sim \mathcal{N}(0,\sigma)$. The effect of noise can be reduced by sampling every condition many times and taking the mean or median, or by running the optimization many times. However, these strategies can be prohibitively time consuming in experimental optimization, where each measurement can take minutes and optimization runs need to converge on the first attempt. Instead, a realistic sampling time must be allocated to each candidate before or during optimization. \changea{This problem has been recognized and explored in simulated optimization, where methods like Optimal Computing Budget Allocation  aim to allocate samples to alternatives with the highest expected value of information \citep{xiao2014simulation, Bartz-Beielstein2007}.} When allocating sample time, each condition can be time-consuming and accurate, or fast and inaccurate; over an optimization run the former strategy takes longer to converge but arrives at a better optimum, while the latter converges faster but with poorer optimum quality \citep{aizawa_scheduling_1994}.  In real-world experimental optimization, the balance between speed and accuracy is often made simply by choosing a single, moderate sample time and assessing every condition with the same precision \citep{ding_human---loop_2018, zhang2017human}. \changea{In this paper, when referring to the amount of sampling budget allocated to a candidate, we use the phrase ``sample time," which could mean either a continuous amount of time or a discrete amount of samples taken of the same candidate. For the example case of exoskeleton optimization, sample time refers to the amount of minutes that a single assistance condition would be tested for as the participant approaches steady state.} \changeb{Sample time could also be referred to as a discrete number of samples (e.g. a 30 second condition could also be measured as 10 breath samples at 3 seconds per breath); we use sample time in this paper to align with common practice in biomechanics research, where treadmill walking trials are defined by time.}

Covariance matrix adaptation evolution strategy (CMA-ES) \citep{hansen_reducing_2003} is a stochastic evolutionary algorithm that seeks to minimize the black-box objective function $f : x \in \mathbb{R}^n \to \mathbb{R}$. CMA-ES operates by sampling a generation of candidates based on a covariance matrix, evaluating each candidate's fitness (i.e. cost), selecting the top performers, updating the covariance matrix accordingly, and then repeating the process. CMA-ES is well suited for experimental robot optimization because it does not require derivative information, and is robust in cost functions with noise, nonconvexity, parameter interaction, and multi-modality \citep{hansen2016cma, groves_sequential_2018, kochenderfer2019algorithms}. CMA-ES has applications in experimental robot optimization, notably in exoskeleton assistance optimization. The use of CMA-ES in human-in-the-loop optimization of ankle exoskeleton assistance improved the field’s largest reduction of energy cost of walking from 14\% to 24\% \citep{zhang2017human}; subsequent exoskeleton studies using CMA-ES have demonstrated the algorithm’s efficacy in high-dimensional spaces \citep{franks_comparing_2021}, in multi-objective optimization \citep{lakmazaheri2024optimizing}, and outside the laboratory \citep{slade_personalizing_2022}. Examples of experimental CMA-ES beyond exoskeleton optimization include optimization of robotic grasping \citep{hu_evolution_2019}, robot movement control and path planning \citep{hill_online_2020, shafii_learning_2015, sharifzadeh_maneuverable_2021}, combustion feedback control \citep{hansen_method_2008}, and obstacle detection \citep{bergener_parameter_2001}. 

An alternative to CMA-ES is Bayesian optimization, which is a sample-efficient optimization method for expensive or noisy black-box functions. \changea{It builds a probabilistic surrogate model of the cost landscape using kriging, continuously refines its predictions and uncertainty estimates as new samples are made. These samples are chosen using an acquisition function, which balances exploration of uncertain regions with exploitation of areas likely to yield better performance  \citep{forrester2008engineering, kochenderfer2019algorithms}.} Bayesian optimization has also demonstrated success in human-in-the-loop exoskeleton optimization \citep{ding_human---loop_2018}, and is believed to have an advantage over CMA-ES in lower-dimensionality search spaces with approximately 20 or fewer parameters \citep{kutulakos2024simulating}. However, the effects of sampling noise, local minima frequency, and parameter interactions on the relative performance of these two optimization algorithms are unclear. 

Special variants of evolutionary algorithms have been proposed for use in environments with high noise \citep{rakshit_noisy_2017}:

\textit{Explicit} and \textit{implicit averaging} \citep{di_pietro_applying_2004, arnold_comparison_2003} use resampling to achieve a predetermined threshold of precision. While explicit averaging samples the same candidate multiple times and implicit averaging samples more candidates per generation, both seek to reduce the amplitude of noise affecting the optimizer. However, this approach can lead to redundant sampling, which is time-consuming and can reduce efficiency.

\textit{Model fitting} \citep{branke_efficient_2001} and \textit{spatial averaging} \citep{sano_optimization_2002} both seek to smooth the cost landscape over the parameter space using prior measurements. These methods work by using previously sampled points to create a local model or to create pseudo-samples to inform the cost of the candidate of interest. While these strategies are beneficial because they enable the leveraging of prior knowledge, they may perform poorly in sparsely sampled regions of the cost landscape. 

\textit{Parameter adapting} variants of CMA-ES \citep{nomura_cmaes_2024, nishida2018psa} have recently been proposed; these algorithms modify the learning rate or population size based on an estimate of the confidence in the update of the CMA-ES distribution parameters. While these techniques demonstrate promising utility, they do not address the challenge of determining how much sampling should be allocated to each candidate. 

Lastly, \textit{dynamic resampling techniques} \changea{\citep{groves_sequential_2018, heidrich2009hoeffding, hansen_method_2008, schmidt2006integrating}} create an online estimate of \changeb{confidence in the current candidates' fitness rank order} by sampling all candidates in a generation, then allocating more samples to selected candidates. \changea{A strength of such approaches is that the amount of sampling per candidate can be} \changeb{adjusted based on previously-measured fitness values for the candidates of that generation.} \changea{While this within-generation sampling allocation is beneficial, these approaches do not specify a method for adapting the per-generation sampling budget, instead opting to fix this value before optimization begins. However, this may be limiting in scenarios where it could be advantageous to sample one generation more than another, for example, if that generation is in a more shallow part of the cost landscape where candidates will have more similar costs.} Additionally, sampling candidates multiple times to discern \changeb{the amount of additional samples (if any) that are required} may lead to unnecessary sampling. Lastly, these sampling strategies may result in ``doubling back," \changeb{where candidates that have been previously evaluated are evaluated again}. This could be undesirable in situations where switching between candidates should be minimized. \changea{For example, in dexterous robotic manipulation tasks, evaluating a new control strategy may require physical reconfiguration of the workspace, reinitialization of sensor baselines, or regrasping of objects—each of which incurs substantial overhead \citep{luo2024precise}. This makes frequent switching between candidates inefficient and disruptive to experimental throughput. In exoskeleton assistance optimization, it takes minutes for a human's cardiorespiratory system to reach steady-state every time exoskeleton assistance changes \citep{selinger2014estimating}. This means that each additional change in assistance requires additional experiment time, as the participant needs time to come to steady state again.} \changea{While algorithms that aim to more efficiently allocate samples to a generation of candidates by avoiding doubling back have been developed for particle swarm optimization \citep{Bartz-Beielstein2007} and simulated annealing \citep{schutte2024improving}, work towards such aims with CMA-ES is limited. }

Building on these approaches, our \changea{research goal was to develop a modification to CMA-ES, called Adaptive Sampling CMA-ES (AS-CMA), to make experimental optimization in the presence of noise faster, more efficient, and more reliable. AS-CMA includes the following key features in support of these goals:} First, AS-CMA enables the use of basic prior knowledge about the sampling method and cost landscape that could be acquired without doing an optimization run. This includes knowledge of how the number of samples of a candidate affects that candidate's noise level. Second, AS-CMA does not re-measure candidates that have already been measured. Third, AS-CMA does not rely on candidates that were sampled in much earlier iterations, which maintains the advantage of CMA-ES being mostly agnostic to samples before the current generation. \changea{This property is valuable when optimizing in time-varying contexts, such as human-robot systems, where the human's cost landscape may evolve as they progress from novice to expert \citep{poggensee_how_2021}. Finally, AS-CMA allows sampling time to vary across generations, enabling greater precision when it is needed, and faster iteration when it is not. }

\changea{The remainder of this paper is structured as follows: in Section 2, we describe CMA-ES, our proposed AS-CMA method, and an alternative approach using Bayesian optimization, then define our simulated optimization framework and outcome metrics. Section 2 concludes by detailing a laboratory experiment where AS-CMA is used to optimize ankle exoskeleton assistance with the objective of minimizing the energy cost of walking. In Section 3, we assess how adaptive sampling affects the sorting stage of CMA-ES, and evaluate the simulated optimization outcomes from CMA-ES, AS-CMA, and Bayesian optimization by comparing the algorithms' convergence speed, efficiency, and reliability. We also give results from the laboratory exoskeleton experiment by showing the movement of selected parameters and resultant costs over the course of the experiment. Section 4 contextualizes our results by summarizing the benefits and potential limitations of adaptive sampling, as well as considering the potential implications of these findings on future real-world optimization applications. Finally, Section 5 concludes the paper by summarizing our findings, suggesting future work, and sharing a link to our open-source code repository. 
}
\section{Methods}
\subsection{CMA-ES}
CMA-ES \citep{hansen2016cma} is a black-box evolutionary optimization algorithm that operates by iteratively selecting candidate points from a multivariate normal distribution, evaluating them, and then updating that distribution based on the best-performing candidates. Here, we separate the algorithm into four steps. 

\textbf{Candidate creation step.} First, $\lambda$ candidates are sampled from $\bm{\mathcal{N}}(\bm{m}, \sigma^2\bm{C})$, where mean $\bm{m} \in \mathbb{R}^n$ defines the center of the distribution, covariance matrix $\bm{C} \in \mathbb{R}^{n \times n}$ defines the shape of the distribution, and step size $\sigma$ sets the spread of the distribution from which candidates are drawn. Each candidate $\bm{x}_i$ is found in generation $g$ as:
\begin{equation}
\bm{x}_i \sim \textbf{m}_g + \sigma_g \bm{\mathcal{N}}(\bm{0}, \bm{C}_g) \quad \text{for } i = 1, \ldots, \lambda
\label{sampling}
\end{equation}
$\lambda$ is typically set to $4 + \lfloor 3 \ln(n) \rfloor$, where $n$ is the dimensionality of the optimization problem \citep{hansen2016cma}.  

\textbf{Evaluation step.} Each candidate $\bm{x}_i$ is evaluated, which determines the fitness $y_i$ of that candidate. An evaluation may involve a single sample or multiple samples that are synthesized into a single value by averaging or fitting a model. 

\textbf{Selection step.} After determining $y_i$ for every candidate, candidates are sorted based on fitness, producing a sorted array $\bm{x}_{i:\lambda}$, where the subscript $i$:$\lambda$ corresponds to the $i$-th best candidate. 

\textbf{Update step.} The multivariate normal distribution state variables $\bm{m}$, $\sigma$, and $\bm{C}$ are updated based on $\bm{x}_{i:\lambda}$. $\bm{m}$ is updated from $\Delta \bm{m}$ from a weighted sum of the top $\mu$ candidates:
\begin{equation}
    \Delta \bm{m} = \sum_{i=1}^{\mu} w_i (\bm{x}_{i:\lambda} - \bm{m}_g)
    \label{eq:mean_update}
\end{equation}
\begin{equation}
    \bm{m}_{g+1} = \bm{m}_{g}+ \Delta \bm{m}
\end{equation}
where $\mu$ is often set to $\lfloor \lambda/2 \rfloor$ and weights $w_1 \geq w_2 \geq \dots \geq w_\mu \geq 0$ sum to 1 \citep{hansen2016cma}. This update shifts the mean towards high-performing candidates from the most recent generation. 

Updating step size $\sigma$ involves first updating the conjugate evolution path $p_\sigma$ as
\begin{align}
    \bm{p}_{\sigma, g+1} &= (1-c_\sigma) \bm{p}_{\sigma, g}  \nonumber
        \\&+ \sqrt{c_\sigma(2-c_\sigma) \mu_{eff}} (C_g)^{-\frac{1}{2}} \frac{\bm{m}_{g+1} - \bm{m}_g}{\sigma_g}
\end{align}
where $c_\sigma^{-1} \approx n/3$ and $\mu_{eff} = (\sum_{i=1}^{\mu} w_i^2)^{-1}$. Then the step size $\sigma$ is updated as 
\begin{equation}
    \sigma_{g+1} = \sigma_g \exp \left(\frac{c_\sigma}{d_\sigma} \left(\frac{\|\bm{p}_{\sigma, g+1}\|}{\mathbb{E}\|\mathcal{\bm{N}}(\bm{0},\bm{I})\|} - 1\right)\right)
\end{equation}
where $d_\sigma \approx 1$ is the damping parameter. This update, known as cumulative step length adaptation, increases $\sigma$ if $\Delta \bm{m}$ points in a similar direction in multiple consecutive updates, and decreases $\sigma$ if $\Delta \bm{m}$ is anti-correlated in consecutive updates.    

Updating the covariance matrix begins with calculation of the evolution path $p_c$ as 
\begin{align}
    \bm{p}_{c, g+1} &= (1-c_c) \bm{p}_{c, g} \nonumber \\
        & + h_\sigma \sqrt{c_c(2-c_c) \mu_{eff}} \frac{\bm{m}_{g+1} - \bm{m}_g}{\sigma_g}
\end{align}
where $c_c^{-1} \approx n/4$ and $h_\sigma$ is 1 if $||p_\sigma|| \lesssim 1.5 \sqrt{n}$, or otherwise 0. Then, $\bm{C}$ is updated as
\begin{align}
    \bm{C}_{g+1} &= (1 + c_1 \Delta(h_\sigma) - c_1 - c_\mu) C_g + c_1 p_c p_c^\intercal \nonumber \\
     & + c_\mu \sum_{i=1}^\mu w_i 
    \frac{\bm{x}_{i:\lambda} - \bm{m}_g}{\sigma_g} \left( \frac{\bm{x}_{i:\lambda} - \bm{m}_g}{\sigma_g}\right )^\intercal
\label{eq:cov_update}
\end{align}
where $c_1 \approx 2/n^2$, $c_\mu \approx \mu_{eff}/n^2$, and $\Delta(h_\sigma) = (1-h_\sigma)c_c(2-c_c) \leq 1$. In essence, the covariance matrix update results in the covariance matrix elongating in the approximate direction of the landscape's gradient and contracting in directions orthogonal to the gradient.

\subsection{Adaptive Sampling Overview}

\begin{figure}[t!]
    \centering
    \includegraphics[width=.8\linewidth]{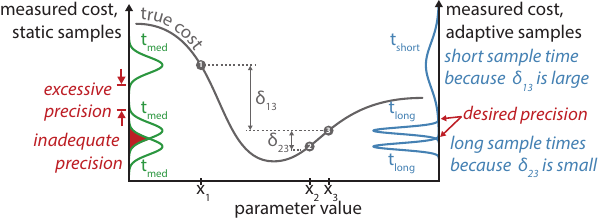}
    \caption{Overview of sample allocation problem posed by noisy optimization and the proposed solution in adaptive sampling. If all candidates are measured for a medium amount of time \changea{(probability distribution for these measurements shown in green)}, excessive or inadequate sorting precision may result. If measurement time instead were based on the difference between the candidate and its nearest-costing neighbor $\delta$, an appropriate amount of sample time \changea{(probability distribution for these measurements shown in blue)} could be allocated.}
    \label{fig:summary_problem}
\end{figure}

AS-CMA is a systematic method for allocating \changea{sample time $t_i$ to candidates $i=1\dots \lambda$ }chosen in the candidate creation step. The way sample time is allocated affects optimization efficiency: more samples of a candidate improves its cost estimation accuracy, but excessive sampling may waste time that would be better spent evaluating new candidates. If insufficient time is spent sampling the candidate, then its fitness is likely to be over- or under-estimated, which may adversely affect optimization (Figure \ref{fig:summary_problem}). 

\begin{figure*}[t!]
    \centering
    \includegraphics[width=1\linewidth]{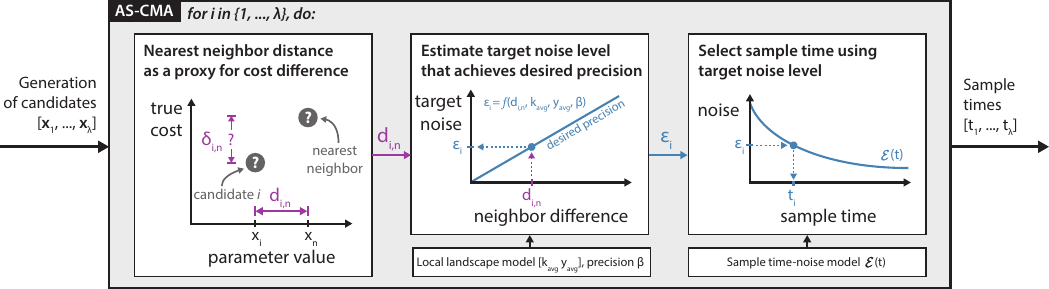}
    \caption{Summary of how AS-CMA determines sample time. For each candidate $i$, the estimated cost difference to its nearest neighbor $\delta_{\text{nearest}}^{(i)}$ is unknown prior to sampling, so AS-CMA uses the nearest neighboring parameter's distance $d^{(i)}_{\text{nearest}}$ instead (Algorithm \ref{alg: AS-CMA}, line \ref{line:nn_dist}). Using $d^{(i)}_{\text{nearest}}$, the target noise level $\epsilon_i$ is found by using an approximation of the local cost landscape ($k_{avg}$ and $y_{avg}$) such that the sorting precision of the candidate and its nearest neighbor should achieve the desired precision $\beta$. (Algorithm \ref{alg: AS-CMA}, line \ref{line:target_noise}). Using $\epsilon_i$, sample time $t_i$ can be found by inverting noise model $\mathcal{E}(t)$ (Algorithm \ref{alg: AS-CMA}, line \ref{line:meas_ti}). }
    \label{fig:overview_ascma}
\end{figure*}

AS-CMA leverages the fact that the selection stage of CMA-ES uses the rank ordering of candidates, not the actual measured values, meaning the amount of noise a candidate can tolerate is related to how distinct its fitness is. If a candidate's $i$'s cost difference to its most similar cost neighbor, $\delta_{\text{nearest}}^{(i)}$, is large, that candidate can tolerate more noise and thus needs fewer samples. Conversely, if two candidates in a generation have similar fitness, they will need to be evaluated with a greater number of samples to ensure proper sorting accuracy in the selection step (Figure \ref{fig:summary_problem}). By allocating sample time based on $\delta_{\text{nearest}}^{(i)}$, AS-CMA seeks to achieve consistent signal-to-noise ratio, where signal is $\delta_{\text{nearest}}^{(i)}$ and noise is related to sample time. In practice, $\delta_{\text{nearest}}^{(i)}$ is not known before candidates are evaluated, so a proxy, $\hat{\delta}^{(i)}_{\text{nearest}}$, is needed. AS-CMA estimates $\hat{\delta}^{(i)}_{\text{nearest}}$ using the distance $d^{(i)}_{\text{nearest}}$ between candidate $i$ and its nearest neighbor (Figure \ref{fig:overview_ascma}), reasoning that candidates that are closer together will be more difficult to sort. This distance is scaled by an estimate of the local slope, $k_{avg}$, to get $\hat{\delta}^{(i)}_{\text{nearest}}$. This scaling accounts for the fact that closely grouped candidates on a steep region of the landscape will be easier to sort than if they were in a shallow region. 

Knowing $\hat{\delta}^{(i)}_{\text{nearest}}$, we next determine how much noise that candidate can tolerate. We denote this noise $\epsilon_i = \mathcal{E}(t_i)$, where $\epsilon_i$ is the magnitude of the standard deviation of the error distribution (with a mean centered at 0) when sampling the candidate for time $t_i$ with noise model $\mathcal{E}(t)$. We model noise as being a percentage of the true cost, i.e. percent error, for consistency across different cost landscapes.  The noise when measuring candidate $i$ affects the measured cost difference between the candidate and its nearest neighbor. The noise in this inter-candidate difference is denoted $\sigma^{(i)}_{\text{nearest}}$ and is proportional to $\epsilon_i$. Together, $\hat{\delta}^{(i)}_{\text{nearest}}$ and $\sigma^{(i)}_{\text{nearest}}$ parameterize the ``estimated detected difference'' distribution, with a mean $\hat{\delta}^{(i)}_{\text{nearest}}$ and standard deviation $\sigma^{(i)}_{\text{nearest}}$. In this distribution, the area above zero is equal to the likelihood of correctly sorting candidate $i$ and its nearest neighbor. Finally, we can choose $t_i$ to ensure that the signal-to-noise ratio, $\beta =\hat{\delta}^{(i)}_{\text{nearest}}/\sigma^{(i)}_{\text{nearest}}$, is large enough to achieve precise sorting, while also small enough to prevent wasted samples. 

We expected that AS-CMA would generally increase sampling time over an optimization run because, as the optimizer approached a point of minima, neighbor distance $d^{(i)}_{\text{nearest}}$ and steepness $k_{avg}$ would decrease. We hypothesized that this effect would improve optimization by allowing for rapid evaluation of diverse candidates early in optimization and precise evaluation of more promising and similar candidates selected later in optimization. Furthermore, sample time selection can also change to match the landscape's local characteristics without requiring any metaparameter adjustments. For example, if the optimizer moved from a shallow to a steep region of the cost landscape, $d^{(i)}_{\text{nearest}}$ and $k_{avg}$ would increase, decreasing sample time so that the optimizer can quickly move down the gradient.

\subsection{Adaptive Sampling Implementation}

\textbf{Inputs and Initialization.} \changea{Following CMA-ES initialization}, AS-CMA requires initialization of three \changea{additional} state variables, $d_{max}$, $k_{avg}$, and $y_{avg}$, which are used for sample time calculations \changea{in every generation  and updated as new measurements are made}. These variables are initialized for the first CMA-ES generation using four values: the maximum and minimum allowable candidate values in the search space $\bm{x}_{max}$ and $\bm{x}_{min}$ \changea{(which are already defined in CMA-ES initialization)}, and \changea{user-input} estimates of the maximum and minimum costs that could be measured in the search space $\hat{y}_{max}$ and $\hat{y}_{min}$. A high level of accuracy in estimating $\hat{y}_{max}$ and $\hat{y}_{min}$ is not necessary for the success of AS-CMA, as the updates done to the state variables after the first generation will use measured values instead of $\hat{y}_{max}$ and $\hat{y}_{min}$, which will allow for the algorithm to adapt to the shape of the local region of search. 

The maximum possible candidate separation, $d_{max}$, is held constant throughout optimization, and is found as the 2-norm distance between $\bm{x}_{max}$ and $\bm{x}_{min}$:
\begin{equation}
    d_{max} = \|\bm{x}_{max} - \bm{x}_{min}\|_2
    \label{eq:d_max}
\end{equation}

The average cost of the local region of search,  $y_{avg}$, is initialized by averaging $\hat{y}_{max}$ and $\hat{y}_{min}$ as

\begin{equation}
    y_{avg} = \frac{\hat{y}_{max} + \hat{y}_{min}}{2}
    \label{eq:y_avg}
\end{equation}
The distance-difference relationship of the local region of search, $k_{avg}$, will be used to give an approximation of how different two candidates' costs are given their distance apart. $k_{avg}$ is initialized before the first generation using $\hat{y}_{max}$ and $\hat{y}_{min}$, assuming that the distance between the largest and smallest possible values in the cost landscape is separated by half of the maximum possible distance:
\begin{equation}
    k_{avg} = \frac{\hat{y}_{max} - \hat{y}_{min}}{0.5 \cdot d_{max}}
    \label{eq:k_avg}
\end{equation}

In addition to $\hat{y}_{max}$ and $\hat{y}_{min}$, two more inputs are required before AS-CMA can begin. These are $\beta$, which is the desired signal-to-noise ratio when sorting candidates, and $\mathcal{E}(t)$, which is the function mapping sample time to the magnitude of the standard deviation of the expected error distribution with sample time $t$. The use of $\beta = 1.3$ is recommended, as it gives empirically strong performance across a range of different cost landscapes with varying dimensionality. \changea{If $\beta$ were increased, AS-CMA would favor longer sample times, which would allow for a higher-quality solution by the conclusion of optimization at the cost of taking more time to converge. The opposite outcome would occur if $\beta$ were made smaller.} $\mathcal{E}(t)$ could be a continuous function, such as a decreasing exponential function or a piecewise function interpolated from experimental data.

\textbf{Assignment of sample time.} The primary function of AS-CMA is to select an appropriate amount of sampling time for each candidate in a generation. This occurs after the candidate creation step and before the evaluation step. After $\lambda$ candidates are chosen as in Eq. \ref{sampling}, the normalized distance from candidate $i$ to its nearest neighbor $d^{(i)}_{\text{nearest}}$is calculated as 
\begin{equation}
    \bm{d}^{(i)}_{\text{nearest}} = \min_{1 \leq j \leq \lambda, j \neq i} \left( \frac {\left[ \| \bm{x}_i - \bm{x}_j \|_2 \right]} {d_{max}} \right)
\end{equation}
Knowing $d^{(i)}_{\text{nearest}}$, the estimated cost difference between candidate $i$ and its nearest neighbor, $\hat{\delta}^{(i)}_{\text{nearest}}$, is found as
\begin{equation}
    \hat{\delta}^{(i)}_{\text{nearest}} = k_{avg} \cdot d^{(i)}_{\text{nearest}}
    \label{eq:delta_hat}
\end{equation}

Using this estimate of $\hat{\delta}^{(i)}_{\text{nearest}}$, the amount of acceptable noise can be calculated to achieve the desired signal-to-noise ratio. To estimate the magnitude of measurement noise, we temporarily assume that candidate $i$ and its nearest neighbor's costs are equal to the average cost of the local region of search $y_{avg}$, and that these two measurements are subject to the same amount of noise, $\epsilon_i$ (in units of percent). Then, the standard deviation of the measured costs for the candidate and its nearest neighbor, $\sigma_i$ and $\sigma_{\text{n}}$, are calculated as 
\begin{equation}
    \sigma_i = \sigma_{\text{n}} = y_{avg} \cdot \epsilon_i
    \label{eq:candidate_noise}
\end{equation}

$\sigma_i$ and $\sigma_{\text{n}}$ are directly related to the standard deviation of the measured difference between candidates $\sigma^{(i)}_{\text{nearest}}$ as \citep{branke2003selection}:
\begin{equation}
    \sigma^{(i)}_{\text{nearest}} = \sqrt{\sigma_i^2 + \sigma_n^2}
    \label{eq:difference_noise}
\end{equation}

Combining equations \ref{eq:candidate_noise} and \ref{eq:difference_noise} gives the standard deviation of the measured difference between the candidates $\sigma^{(i)}_{\text{nearest}}$ as a function of each candidate's measurement noise:
\begin{equation}
    \sigma^{(i)}_{\text{nearest}} = \sqrt{2} \cdot y_{avg}  \cdot \epsilon_i
    \label{eq:difference_noise_epsilon}
\end{equation}

In essence, $\sigma^{(i)}_{\text{nearest}}$ is the noise involved in the sorting task between the candidate and its nearest neighbor. For a candidate with a larger estimated cost difference between itself and its nearest neighbor, $\hat{\delta}^{(i)}_{\text{nearest}}$, a larger value of $\sigma^{(i)}_{\text{nearest}}$ is acceptable. Conversely, a candidate with a very similar cost to its neighbor, i.e. a small $\hat{\delta}^{(i)}_{\text{nearest}}$, will require greater precision in its measurement, meaning a smaller $\sigma^{(i)}_{\text{nearest}}$ is needed. 

The desired signal-to-noise ratio metaparameter is found as $\beta = \hat{\delta}^{(i)}_{\text{nearest}} / \sigma^{(i)}_{\text{nearest}}$. Substituting for $\sigma^{(i)}_{\text{nearest}}$ using equation \ref{eq:difference_noise_epsilon} and for $\hat{\delta}^{(i)}_{\text{nearest}}$ using equation \ref{eq:delta_hat}, then solving for $\epsilon_i$ gives:
\begin{equation}
        \epsilon_i = \frac{k_{avg} \cdot d^{(i)}_{\text{nearest}}}{\sqrt{2} \cdot \beta \cdot y_{avg}}
        \label{eq:desired_error}
\end{equation}

Equation \ref{eq:desired_error} summarizes the decision made by AS-CMA. That is, by using knowledge of the local cost landscape ($k_{avg}$ and $y_{avg}$), AS-CMA identifies the maximum amount of noise $\epsilon_i$ that can be tolerated while still allowing correct sorting of a neighbor of distance $d^{(i)}_{\text{nearest}}$ with precision parameterized by $\beta$. 

After finding $\epsilon_i$, the sample time $t_i$ can be found by inverting the model mapping measurement time to measurement noise $\mathcal{E}(t) = \epsilon$. This sample time calculation process is repeated for all candidates $i = 1, \dots, \lambda$ in the generation. Then, candidates $\bm{x}_{1:\lambda}$ are evaluated for time $t_{1:\lambda}$ to identify costs $y_{1:\lambda}$. Then the CMA-ES selection and update steps occur. 

\begin{algorithm}[t!]
  \caption{Adaptive Sampling CMA-ES}
  \begin{algorithmic}[1]
    \State \textbf{input AS-CMA Parameters:} $\hat{y}_{max}$, $\hat{y}_{min}$, $\beta$, $\mathcal{E}(t)$   
    \State \textbf{initialize:} $d_{max}$, $y_{avg}$ $k_{avg}$ per eq. \ref{eq:d_max}-\ref{eq:k_avg} 
    
    \While{\text{not terminated}}
        \LComment{CMA-ES: Candidate creation}
        \For{$i$ in $\{1, \ldots, \lambda\}$}
            \State $\bm{x}_i \sim \textbf{m}_g + \sigma_g \bm{\mathcal{N}}(\bm{0}, \bm{C}_g)$ 
        \EndFor
      
        \LComment{AS-CMA: Determine sample time}
        \For{$i$ in $\{1, \ldots, \lambda\}$}
            \State $d^{(i)}_{\text{nearest }} = \min_{1 \leq j \leq \lambda, j \neq i} \left( \frac{\left[ \| \bm{x}_i - \bm{x}_j \|_2 \right]}{d_{max}} \right)$ \label{line:nn_dist}
            \State $\epsilon_i \leftarrow \frac{k_{avg} \cdot d^{(i)}_{\text{nearest}}}{\sqrt{2} \cdot \beta \cdot y_{avg}}$ \label{line:target_noise} 
            \State  $t_i \leftarrow \underset{t}{\mathrm{arg\min}} |\mathcal{E}(t) - \epsilon_i|$ \label{line:meas_ti}
        \EndFor
        
        \LComment{CMA-ES: Evaluation, selection, update}
        \For{$i$ in $\{1, \ldots, \lambda\}$}
        \State $y_i \leftarrow f_{meas}(\bm{x}_i, t_i)$
        \EndFor
        \State $\bm{x}_{i:\lambda} \leftarrow \textrm{sort}(\bm{x}, \bm{y})$
        \State Update $\bm{m}$, $\sigma$, $\bm{C}$ per eq. \ref{eq:mean_update}-\ref{eq:cov_update}

        \LComment{AS-CMA: Update}
        \State $y_{avg} \leftarrow \frac{1}{\lambda} \sum_{i=1}^{\lambda} y_i$
        \State $\bm{d}_y = \left[ (y_i - y_j) \right]_{1 \leq i, j \leq \lambda}$
        \State $\bm{d}_x = \left[ \| \bm{x}_i - \bm{x}_j \|_2 \right]_{1 \leq i, j \leq \lambda}$
        \State  $k_{avg} = \underset{k}{\mathrm{arg\min}}  \| \bm{d}_y - k \bm{d}_x \|_2 ^2$
        \EndWhile
  \end{algorithmic}
  \label{alg: AS-CMA}
\end{algorithm}

\textbf{Update of local cost landscape estimation.} 
After the CMA-ES update step, AS-CMA state variables $y_{avg}$ and $k_{avg}$ are updated to inform cost difference estimation in the next generation. Updating $y_{avg}$ is trivial:
\begin{equation}
    y_{avg} = \frac{1}{\lambda} \sum_{i=1}^{\lambda} y_i
\end{equation}
Updating $k_{avg}$ requires fitting a distance-difference linear model to data from the current generation. First, vectors made up of the pairwise candidate cost differences $\mathbf{d}_y = \left[ (y_i - y_j) \right]_{1 \leq i, j \leq \lambda}$ and parameter distances $\mathbf{d}_x = \left[ \| \mathbf{x}_i - \mathbf{x}_j \|_2 \right]_{1 \leq i, j \leq \lambda}$ are calculated. Then, the cost difference-parameter distance relationship between each candidate and each of its neighbors is fit using least squares as
\begin{equation}
    k_{avg} = \underset{k}{\mathrm{arg\min}}  \| \bm{d}_y - k \bm{d}_x \|_2 ^2
\end{equation}
After $y_{avg}$ and $k_{avg}$ have been updated, the next generation can begin with the candidate creation step. 

\changeb{AS-CMA maintains the same order of computational complexity as CMA-ES. The main additional cost arises from calculating pairwise candidate distances within each generation, which scales as $O(\lambda^2)$. Updates to $k_{avg}$ and $y_{avg}$ involve simple vector operations and a least-squares fit, which scale linearly with $\lambda$ and the problem dimension $n$. }A summary of AS-CMA in the context of CMA-ES is given in Algorithm \ref{alg: AS-CMA}. 

\subsection{Evaluation in Simulation}

We compared CMA-ES with various static sample times to AS-CMA in four cost landscapes with features that are relevant to a range of optimization problems. \changea{The number and types of landscapes and the variety of noise levels were chosen to align with similar studies of evolutionary algorithms for noisy optimization \citep{groves_sequential_2018, kutulakos2024simulating, hansen_method_2008}.} The tested landscapes were:

\begin{enumerate}
    \item A 4-dimensional empirical approximation of the metabolic cost landscape of ankle exoskeleton assistance \citep{zhang2017human}, chosen to allow qualitative comparisons to be made between simulated and experimental data. We refer to this landscape as Ankle. 
    \item 4-dimensional generalized Rosenbrock function \citep{rosenbrock1960automatic}, chosen to assess performance in scenarios with interaction between variables. 
    \item 4-dimensional Levy function \citep{laguna_experimental_2005}, chosen to evaluate the effect of many local minima. 
    \item 20-dimensional Sphere function \citep{dixon1978global}, chosen to test how algorithmic performance scales with problem dimension \changea{(e.g. a 22-dimensional exoskeleton optimization problem given in \cite{franks_comparing_2021})}. 
\end{enumerate}

Full equations for these functions are given in Table \ref{tab:functions}, where $n=4$ for Rosenbrock, Levy, and Ankle; $n=20$ for Sphere. \changea{We selected $n=4$ for three of these functions because this problem dimensionality represents a threshold beyond which simpler methods (e.g. grid or random search) are less effective for experimental optimization, and optimization algorithms become increasingly necessary.}

\changea{While the Rosenbrock, Levy, and Sphere functions typically have a minimum of 0 \citep{rosenbrock1960automatic, dixon1978global, laguna_experimental_2005}, we added 100, 10, and 0.67 to these functions so that multiplicative noise would continue to have an effect even as the optimizer approached the minima. 0.67 was chosen for the Sphere function so that its minimum would be of similar magnitude to the Ankles minimum; 100 and 10 were selected for Rosenbrock and Levy to give optimization runs a realistic chance to reach the fine convergence threshold in the presence of the noise we applied}. In $f_{Ankle}$, $x_1$, $x_2$,$x_3$, and $x_4$ represent the exoskeleton peak torque, peak time, rise time, and fall time, respectively, which are described by \citet{zhang2017human}.

\begin{table}[b]
\centering
\caption{Ground-truth test functions used for simulated optimization.}
\label{tab:functions}
\resizebox{\textwidth}{!}{%
\begin{tabular}{l|l}
\hline
Function                                                                                                                & Bounds                                \\ \hline
$f_{Rosenbrock}(\mathbf{x}) = 100 + \sum_{i=1}^{n-1} \left[ 100 \left( x_{i+1} - x_i^2 \right)^2 + (1 - x_i)^2 \right]$ & $x_i \in [-5.12,5.12]$ for $i = 1, \dots, n$ \\
$f_{20D Sphere}(\mathbf{x}) =  0.67 + \sum_{i=1}^{n}x_i^2$& $x_i \in [0,1]$ for $i = 1, \dots, n$ \\
\begin{tabular}[c]{@{}l@{}}$f_{Levy}(\mathbf{x}) = \sin^2(\pi w_1) + \sum_{i=1}^{n-1} (w_i - 1)^2 \left[1 + 10 \sin^2(\pi w_i + 1)\right]$ \\ \hspace{1em}$+ (w_n - 1)^2 \left[1 + \sin^2(2\pi w_n)\right] + 10$, where $w_i = 1+\frac{x_i-1}{4}$ for $i = 1, \dots, n$ \end{tabular} &
  $x_i \in [-10,10]$ for $i = 1, \dots, n$ \\
$f_{Ankle}(\mathbf{x}) = 1 + 0.95 \left( e^{-x_1} - 1 \right) + (x_2 - 1)^2 + 0.1 (x_3 - 0.2)^2 + x_4^2$ &
  \begin{tabular}[c]{@{}l@{}}$x_1 \in [0,1]$, $x_2 \in [0.1,0.55]$, \\ \hspace{1em} $x_3 \in [0.1, 0.4]$, $x_4 \in [0.05, 0.2]$\end{tabular} \\ \hline
\end{tabular}%
}
\end{table}

\begin{figure}
    \centering
    \includegraphics[width=.8\linewidth]{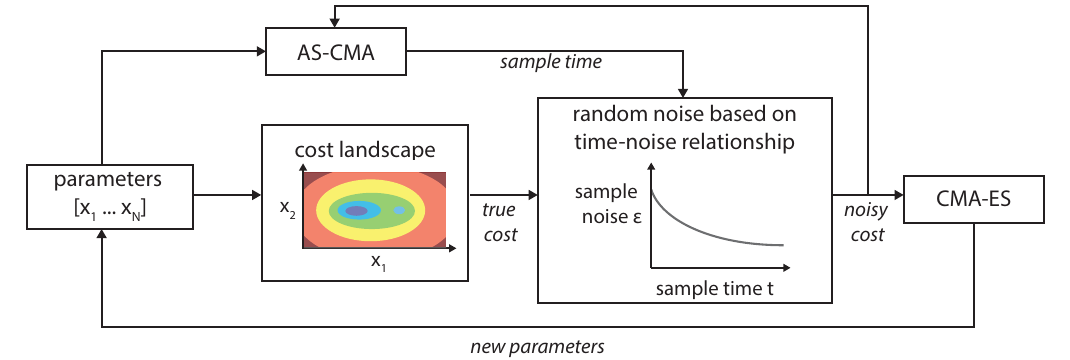}
    \caption{Representation of numerical simulation framework. Given a set of candidate parameters, the simulation first determines the true cost (using a ground-truth cost landscape) and the desired sample time (using AS-CMA, which is shown, or a fixed sample time, not shown). Then, noise is introduced to the true cost based on the chosen sample time, producing the noisy cost. This noisy cost is returned to CMA-ES, which then generates new candidates to assess, and AS-CMA, for updating of $k_{avg}$ and $y_{avg}$. }
    \label{fig:sim_framework}
\end{figure}

During optimization, noise was added based on the selected measurement time $t_i$ (Figure \ref{fig:sim_framework}), \changeb{representing how during actual experiments, the noise from a measurement is dependent on the time allocated to that measurement}. For a given candidate, the true fitness was first calculated (Table \ref{tab:functions}), then noise was introduced as $y_{noisy} = y_{true} \cdot (1 + noise)$, where $noise$ is a random sample from $\mathcal{N}(0, \epsilon_i)$, and $\epsilon_i = \mathcal{E}(t_i)$. \changea{We use multiplicative noise to be consistent with real metabolic data from exoskeleton optimization, where the amount of noise present in energy cost measurements increases with increasing energy cost. }We used an error model $\mathcal{E}(t)$ calculated from data collected in a previous study \citep{zhang2017human}, which consisted of 88 total trials. For this model, percent error for a given sample time was calculated by comparing the estimated metabolic cost for that sample time versus the cost estimated using the entire 6 minute trial. We chose minimum and maximum allowable sample times for simulated optimization of 0.5 and 5.5 minutes, which corresponded to error standard deviations of 34.2\% and 0.4\%. \changea{The minimum sampling time was based on the requirement that human users of exoskeleton assistance require approximately 30 seconds to adapt to assistance; the maximum sampling time was selected based on the time that is considered sufficient to allow a human to come to steady state \cite{slade_personalizing_2022}. Finally, an additional 3\% of baseline noise was added to the noise magnitude to reflect that even a full 6 minute trial is likely not perfectly accurate relative to an even longer trial, where other biomechanical changes such as gait strategy and neuromuscular adaptation could stabilize. 
  }

For each test function, $N=100$ optimization runs were completed for each static sampling time and AS-CMA. \changea{Throughout optimization, the amount of simulated experiment time was tracked based on the time allocated to each sample, which allowed for optimization outcomes to be framed in terms of real-world experiment time. }For all runs, we used the typical CMA-ES framework and parameter values described by \citet{hansen2016cma} and deployed in the \texttt{cmaes} library \citep{nomura_cmaes_2024}, with the exception that the current generation's mean parameter value was inserted as the final candidate of each generation, reflecting established practice in human-in-the-loop optimization \citep{zhang2017human, slade_personalizing_2022}. Initialization for each optimization run was chosen from a random uniform distribution over the space for all cost landscapes except Ankle, which was started at the same, moderate-cost initial value to reflect real-world exoskeleton optimization methodology \citep{poggensee_how_2021}. Initialization was kept consistent between run numbers (i.e. the $n$th run of each sampling strategy started at the same point). All parameter values were scaled to be between 0 and 1 and the initial step size was set to $\sigma=0.3$ as recommended by \citet{hansen2016cma}. \changea{In AS-CMA runs, we set $\hat{y}_{min}$ to 0.6 for the 20D Sphere and Ankle test cases, and to 0 for Rosenbrock and Levy. For $\hat{y}_{max}$, we used 1.3 for 20D Sphere and Ankle, 1000 for Rosenbrock, and 250 for Levy. These values were chosen to be in the appropriate range but intentionally imprecise relative to the ground-truth cost functions, reflecting how an experimenter might provide imperfect AS-CMA settings before running optimization. }

\subsection{Comparison with KL-KG CMA-ES}
\changeb{To evaluate AS-CMA's performance versus dynamic resampling techniques, we conducted optimization using the method proposed in \citet{groves_sequential_2018}, which we will refer to as Kullback Leibler Knowledge Gradient (KL-KG) CMA-ES. This method uses the KG method to minimize the KL divergence between successive CMA-ES sampling distributions, thereby prioritizing candidate evaluations expected to yield the greatest information gain about the optimal region. KL-KG CMA-ES begins by sampling each candidate \( n_0 \) times to estimate initial fitness values and compute the CMA-ES distribution \((\bm{m}, \Sigma)\) at the start of each generation. It then allocates additional samples to selected candidates until the generation’s total sampling budget \( N_{total} \) is exhausted (the total sampling budget is referred to as $N$ in the original manuscript, but we will use $N_{total}$ here for clarity). These candidates are selected based on their expected contribution to improving the CMA-ES distribution: for each candidate $a_i$, KL-KG estimates the probability $P_i$ that an additional sample will change the current elite set $I_t^{\mu}$ based on the candidate's current estimated fitness $S_i$ and the noise model $\sigma$. This noise model must be defined before the start of optimization (which is also a requirement for AS-CMA). The algorithm then computes the corresponding hypothetical update to the CMA-ES parameters $(\bm{m}_i, \Sigma_i)$. The value of sampling each candidate is then quantified as $ 
V_i = P_i \, D_{\mathrm{KL}}\big(\mathcal{N}(\bm{m}_i, \Sigma_i) \,\|\, \mathcal{N}(\bm{m}, \Sigma)\big),
$ where $D_{\mathrm{KL}}(\cdot\|\cdot)$ denotes the KL divergence between the updated and current CMA-ES distributions. The candidate with the highest $V_i$ is sampled next.}

\changeb{For our tests, we conducted $N=100$ optimization runs for each of the four landscapes. As in the CMA-ES tests, these 100 runs were performed for each static duration to provide KL-KG CMA-ES with the best opportunity to demonstrate its strongest performance. We implemented the optimization and candidate selection procedures described in \citet{groves_sequential_2018} and empirically tuned the initial number of samples per candidate $n_0$ and the total sampling budget per generation $N_{total}$, resulting in $n_0=1$ and $N_{total}=20$.}

\subsection{Comparison with Bayesian Optimization}
Static sampling Bayesian optimization was tested to identify comparative strengths and weaknesses between it and AS-CMA during noisy optimization. $N=100$ Bayesian optimization runs were completed for each static sample time for the Ankle and 20D Sphere landscapes, and $N=50$ runs were completed for the Rosenbrock and Levy landscapes due to computing cluster time limitations during the surrogate model fitting step. Bayesian optimization \citep{nogueira2014bayesian} settings were chosen to align with those found most successful by \citet{kutulakos2024simulating} to enable comparison with this work. The Gaussian process, which is used to model the cost landscape with mean $\mu(x)$ and uncertainty $\sigma(x)$ as successive candidates are assessed, was modeled using a Matern kernel \changeb{and a noise model assuming constant-variance (homoscedastic) noise across the cost landscape}. Unlike in the AS-CMA and KL-KG CMA-ES tests, the Bayesian optimizer did not utilize the known multiplicative noise model, which may have put the Bayesian optimizer at a disadvantage when comparing the algorithms' performance. \changea{The acquisition function, which selects the points in the landscape to evaluate, sought to minimize the lower confidence bound, i.e. $\underset{x}{\mathrm{arg\max}} (\mu(x) - \kappa \sigma(x))$  \citep{srinivas2009gaussian}.} The exploration constant $\kappa$, which defines the optimizer's choice to explore novel regions of the cost landscape versus continuing sampling in promising regions, \changeb{was kept at the optimizer package's value of 2.6 \citep{nogueira2014bayesian}, which corresponds approximately to the 99\% confidence interval of a standard normal distribution}. $\kappa$ was decayed as $\kappa \leftarrow 0.93 \cdot \kappa $ after every sample to create a gradual shift from exploration to exploitation as optimization progressed. \changea{The true cost of the best posterior mean derived from noisy optimization was used to compute Bayesian optimization-estimated optimum at each optimization timestep.}

\subsection{Analysis}
To evaluate how sampling strategies influence the accuracy of the CMA-ES sorting step, we used Spearman's rank correlation coefficient (Spearman's $\rho$) \citep{spearman1961proof}. This measure was chosen because it quantifies the strength and direction of the relationship between two ranked variables, thereby reflecting how closely noisy estimated costs' ordering align with the ground truth ordering. A $\rho$ value of 1 corresponds to perfect ordering, while 0 indicates no relationship between the noisy and true orders. Values of $\rho$ closer to 1 are preferred, as a correctly ranked set of candidates should guide the optimizer toward a minimum. However, a $\rho$ of exactly 1 is not essential, as CMA-ES's population-based iterations allow it to tolerate some noise, and achieving perfect ranking accuracy would likely require prolonged measurements for each candidate. Spearman's $\rho$ was calculated for the $\lambda$ candidates in each generation, then averaged across all 100 runs at each time point for each sampling strategy.

To assess convergence during the optimization run, we first defined the true minimum $y^*$ as $y^* = f_{true}(x^*)$, where $x^*$ is the value of the parameters that minimize the ground truth cost landscape $f_{true}$. At each generation $g$ in an optimization run, the true cost of the estimated optimum is defined as  $\hat{y}^* = f_{true}(\hat{x}^*)$, where $\hat{x}^*$ is the CMA-ES mean value parameter value $m_g$. We considered both “coarse” and “fine” convergence thresholds, which is the point where the $\hat{y}^*$ achieved and stayed within 20\% and 5\% of the true minimum $y^*$ for the remainder of the run. \changea{The 20\% threshold was chosen to reflect a minimum viable outcome in human-in-the-loop exoskeleton optimization, based on the observation that generic assistance often achieves approximately 80\% of the benefit of optimized assistance \citep{poggensee_how_2021}, suggesting that surpassing this level indicates a meaningful improvement. The 5\% threshold was selected as a threshold representing practical convergence, where further optimization may yield diminishing returns, particularly given that this value is comparable to the noise floor of metabolic cost measurements \citep{zhang2017human}.  }

For each optimization run, we evaluated the two convergence thresholds using three metrics:

\begin{enumerate}
    \item \textit{Time to convergence}: The time for $\hat{y}^*$ to get to and stay within the coarse or fine threshold of the true minimum value. This metric represents the need to minimize the \textit{time burden} of optimization. \changea{In this paper, time refers to simulated experiment time, not compute time. }
    \item \textit{Cumulative cost to convergence}: The total cost of all tested conditions required to reach convergence. This metric represents the need to minimize the total \textit{cost burden} of optimization. \changea{This metric was motivated by exoskeleton optimization - during optimization, we would like to minimize the total amount of metabolic energy the participant spends on walking across all conditions. Cumulative cost to convergence would allow us to discriminate between two optimization strategies that achieve a threshold at the same time, but have different energetic expenditures across the course of optimization.}
    \item \textit{Convergence reliability}: The percentage of runs that achieve convergence. This metric illustrates the tradeoff between rapid convergence and reliable convergence. This metric represents the need to minimize the \textit{rerun burden}, or the likelihood that an optimization run will not find a quality solution on its first run and will need to be run again. 
\end{enumerate}

\changea{For runs that did not converge, we assigned a time to convergence equal to the end time of optimization, and a cumulative cost to convergence equal to the sum of the true costs of all conditions tested. We chose not to enforce any additional penalties on these metrics for non-converging runs because non-convergence would be made evident by the convergence reliability metric. }

To compare static and adaptive sampling approaches, a determination of the single best static sampling time for CMA-ES and Bayesian optimization for each cost landscape was necessary. The best static sampling time was determined by quantifying each static run's performance in each of the six metrics, normalizing those performance values to each of the six performance scores assigned to AS-CMA, and then summing across the six scores. Any static sampling time that did not achieve fine convergence in at least 90\% of runs was excluded from consideration for best static time. Comparison between convergence time and convergence cost was made using a two-tailed unpaired t-test; all tests used a significance level of $p = 0.05$. 

\subsection{\changea{Laboratory Pilot Exoskeleton Optimization}}
To qualitatively validate the results found in the simulated optimization tests, we conducted an exoskeleton optimization experiment with $N=1$ participant (male, 26 years, 172 cm, 62 kg) using a bilateral tethered ankle exoskeleton \citep{poggensee_how_2021}, with metabolic cost as an objective function (CPET, COSMED). Optimization was conducted following the human-in-the-loop \changea{CMA-ES} optimization protocol \citep{poggensee_how_2021, zhang2017human}, with one modification: instead of using a fixed 2 minute sample time, measurement time for each condition was chosen using AS-CMA from the range of 0.5 to 5.5 minutes. After optimization, the energy cost of every mean candidate was evaluated for 3 minutes twice in a double reversal (ABCCBA) format, and normalized to the energy cost of walking with the exoskeleton applying zero torque. The progression of the mean parameter values and the cost of the mean were used to assess if optimization would approach the expected optimum identified from previous ankle exoskeleton experiments. We expected the optimizer to identify that high peak torque at a late peak time was optimal, and that optimal assistance would achieve approximately a 39\% reduction in energy cost of walking relative to walking with the exoskeleton turned off \citep{poggensee_how_2021}. All experimental protocols were approved by Stanford University IRB (IRB-48749). 

\section{Results}
\begin{figure}[t!]
    \centering
    \includegraphics[width=0.6\linewidth]{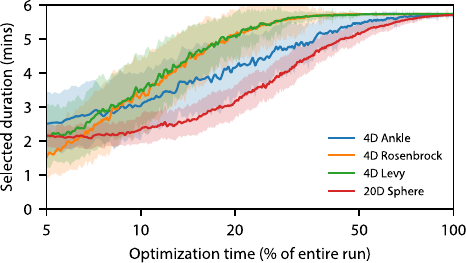}
    \caption{Average ($\pm1$ standard deviation) selected condition time by AS-CMA.}
    \label{fig:sample_vs_exp_time}
\end{figure}

\subsection{AS-CMA selection of sample time}
AS-CMA selected short sample times at the start of optimization runs and generally increased the condition sample time as the run progressed (Figure \ref{fig:sample_vs_exp_time}). This trend was likely a consequence of the optimizer approaching shallower parts of the cost landscape, where candidates were more similar and therefore more difficult to sort. The chosen sample times varied with cost landscape, demonstrating how AS-CMA adapts to different optimization problems. 

\subsection{Sorting accuracy}
\begin{figure*}[t!]
    \centering
    \includegraphics[width=.9\linewidth]{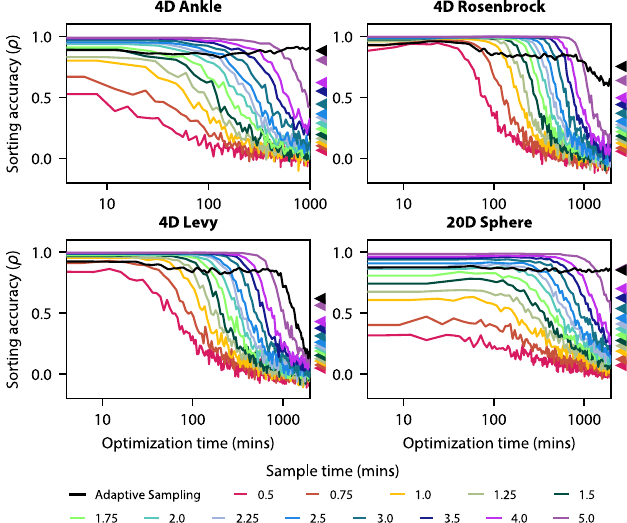}
    \caption{Generation sorting accuracy (Spearman's $\rho$) over optimization runs. Triangles denote the average across all optimization time points for that landscape.}
    \label{fig:sorting_accuracy}
\end{figure*}

Over the course of optimization, sorting accuracy generally decreased for all sampling strategies as the optimizer moved towards shallower regions of the cost landscape (Figure \ref{fig:sorting_accuracy}). AS-CMA had favorable accuracy throughout optimization, starting with $\rho$ between 0.87 and 0.93, and maintaining this accuracy throughout the optimization run (Ankle and 20D Sphere) or losing accuracy at the end of optimization (Rosenbrock and Levy). \changeb{For all test landscapes, average sorting accuracy stayed high further into the optimization run than any of the static durations, indicating the selected sample times for candidates were long enough to achieve appropriate sorting in the CMA-ES selection step. Sorting accuracy remained high throughout the optimization for the Ankle and 20D Sphere functions, while in Rosenbrock and Levy functions exhibited a decline near the end of the run. This falloff is likely due to the optimizer reaching very shallow parts of the cost landscape, where candidates' cost differences were very small relative to the noise. We expect that, had optimization been continued for long enough, AS-CMA would have reached this accuracy falloff point in all of the cost landscapes as it honed in on a minimum point. At that point, even the small amount of error introduced by the longest allowable sample time would make sorting these similar-costing candidates increasingly difficult, resulting in decreased sorting accuracy.} For static time sampling, $\rho$ correlated with sample time: the longer the sampling time, the greater the starting accuracy, and the longer this accuracy was sustained. AS-CMA had the highest $\rho$ for all four landscapes when averaged across all runs and all time points. The point where accuracy began to decline was later for AS-CMA than for all static times, this falloff is likely due to the selected sample time approached the maximum allowable sampling time. The favorable sorting accuracy of AS-CMA in all stages of optimization suggests the algorithm's gradual increase in selected sampling times (Figure \ref{fig:sample_vs_exp_time}) contributed to the sustained high level of sorting accuracy, ultimately making the optimizer more efficient. 

\begin{figure*}[t!]
    \centering
    \includegraphics[width=.9\linewidth]{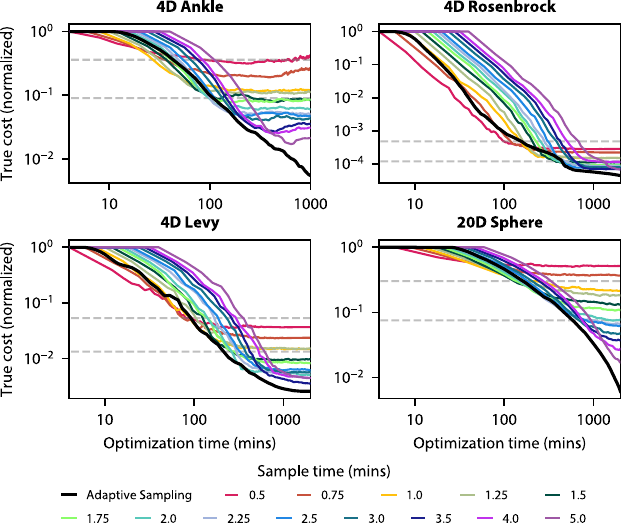}
    \caption{True cost of estimated optimum of CMA-ES runs using either static (colored) or adaptive (black) sampling. Lines represent the average over $N=100$ runs for each sampling strategy. The higher and lower dashed lines indicate the coarse and fine convergence thresholds. Vertical axes are normalized to the cost of the point where optimization was initialized.}
    \label{fig:cma_true_cost_vs_time}
\end{figure*}
\begin{figure*}[t!]
    \centering
    \includegraphics[width=1\linewidth]{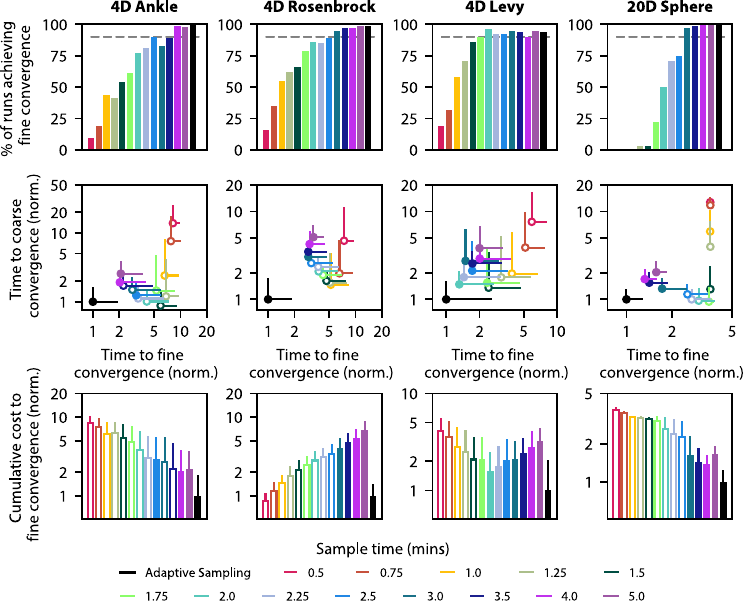}
    \caption{Performance metrics for various static CMA-ES and AS-CMA optimization runs. Row 1: Convergence rate. Static sampling strategies that did not achieve fine convergence on 90\% or more runs (gray dashed line) are indicated in the next two rows with white fill colors. Row 2: Convergence time. Row 3: Convergence costs. Values in the second and third rows are normalized to AS-CMA, where smaller values indicate better performance. Whiskers for the second and third rows represent one standard deviation.}
    \label{fig:cma_metrics}
\end{figure*}

\subsection{Cost of estimated optimum over time}
For all sample times, the true cost of the CMA-ES-estimated optimum asymptotically decreased over time (Figure \ref{fig:cma_true_cost_vs_time}); the rate of this decrease and the value of the asymptote varied with sample time. For short static sample times, cost generally decreased rapidly at the start of optimization, but approached an asymptote that was distant from the true minimum. For longer static sample times, the rate of cost decrease was slower, but functions approached a minimum value that was closer to the global minimum. AS-CMA combined these traits: it quickly reduced cost at the start of optimization (although not as fast as the shortest sample times), yet, at the end of optimization had still found the lowest-costing estimated optimum for all four cost landscapes.  

\subsection{Convergence reliability}
Across 100 runs in each of the four cost landscapes, AS-CMA achieved coarse convergence in all 400 runs, and fine convergence in 393 runs. For static sample times, longer times more reliably reached fine convergence, while the shortest sample times often did not achieve fine convergence (Figure \ref{fig:cma_metrics}, first row). Nearly all sample times reliably attained coarse convergence (Figure \ref{fig:cma_true_cost_vs_time}). \changea{While average cost trajectories for shorter sampling durations may appear to cross the fine convergence threshold earlier (Figure \ref{fig:cma_true_cost_vs_time}), many individual runs fail to remain below the threshold due to high variability in the CMA-ES mean, which results in fewer short-duration runs satisfying the convergence criterion (Figure \ref{fig:cma_metrics}, first row), and therefore larger costs and times to convergence (Figure \ref{fig:cma_metrics}, second and third rows). In contrast, longer sampling times produce more stable cost estimates, allowing a higher proportion of runs to meet and maintain fine convergence. }

\subsection{Static sample time performance}
The overall performance of static CMA-ES sample times formed a Pareto frontier, where shorter sample times achieved coarse objectives, but displayed poor consistency in achieving fine objectives, while long sample times reached both thresholds more slowly (Figure \ref{fig:cma_true_cost_vs_time}). The shortest sample times often did not have 90\% of runs achieve the fine convergence threshold, meaning those sample times were not considered for best static sample time.  Moderate-length sample times returned favorable results in both coarse and fine objectives, although notably, the time of the best static sample time varied depending on the cost landscape. The best \changea{CMA-ES} sample time for Ankle, Rosenbrock, Levy, and 20D Sphere was 4, 3, 2, and 3.5 minutes, respectively. \changea{Percentage comparisons between these best static sampling methods and AS-CMA across the different convergence metrics are given in the Appendix.}

\subsection{Time to convergence}
For all tested cost landscapes, AS-CMA was able to achieve coarse and fine convergence more quickly than all static sampling times (Figure \ref{fig:cma_metrics}, second row).  As compared to this best static sample time for the four landscapes, AS-CMA was 24 to 65\% faster at achieving fine convergence  ($p<0.05$ for Ankle, Rosenbrock, and 20D Sphere, $p=0.14$ for Levy) and 32 to 67\% faster at achieving coarse convergence ($p<0.05$ for all landscapes). AS-CMA seemed to perform much better than CMA-ES in the Ankle and Rosenbrock landscapes, and closer to, but still better than, the best static sampling time in Levy and 20D Sphere. The standard deviation of the time to convergence for AS-CMA was also notably smaller than that of static times, suggesting a more consistent overall performance by the adaptive sampling approach. 

\subsection{Cumulative cost to convergence}
Short sample times had favorable costs to coarse convergence, but poor costs to fine convergence, while longer times had higher relative costs to coarse convergence and lower costs to fine convergence. When the sampling strategy converged quickly, it also did so with lower cumulative cost, and vice versa (Figure \ref{fig:cma_metrics}, third row). This is explained by the plot of mean cost over time (Figure \ref{fig:cma_true_cost_vs_time}), which indicates that the shape of the cost over time is approximately an exponential decay shape across all sampling strategies. As compared to the best static sampling time on each landscape, AS-CMA converged to the fine threshold 29 to 76\% faster, and to the coarse threshold 36 to 76\% faster ($p<0.05$ for all landscapes and thresholds). 

\begin{figure*}[t!]
    \centering
    \includegraphics[width=.9\linewidth]{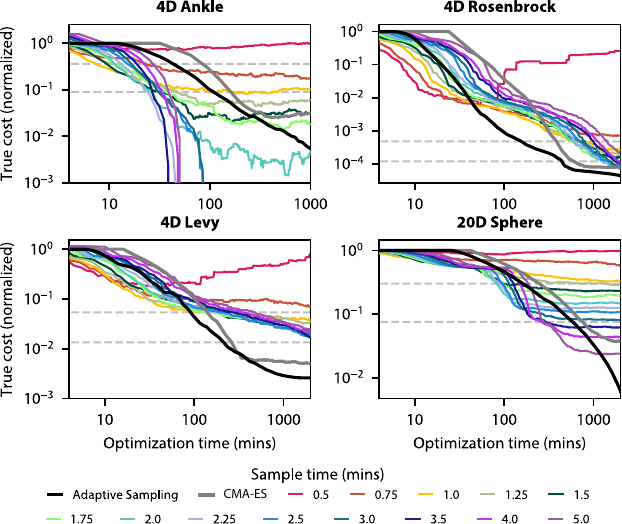}
    \caption{True cost of estimated optimum over Bayesian optimization comparison experiment ($N=100$ runs for all except Rosenbrock and Levy static Bayesian optimization runs, which were $N=50$). The best CMA-ES static sample time (gray line) was 4, 3, 2, and 3.5 minutes for Ankle, Rosenbrock, Levy, and 20D Sphere, respectively. The higher and lower dashed lines indicate the coarse and fine convergence thresholds.  The vertical axes are normalized to the cost of the AS-CMA and CMA-ES optimization starting points, which were the same across sampling strategies.}
    \label{fig:bo_true_cost_vs_time}
\end{figure*}

\begin{figure*}[t!]
    \centering
    \includegraphics[width=1\linewidth]{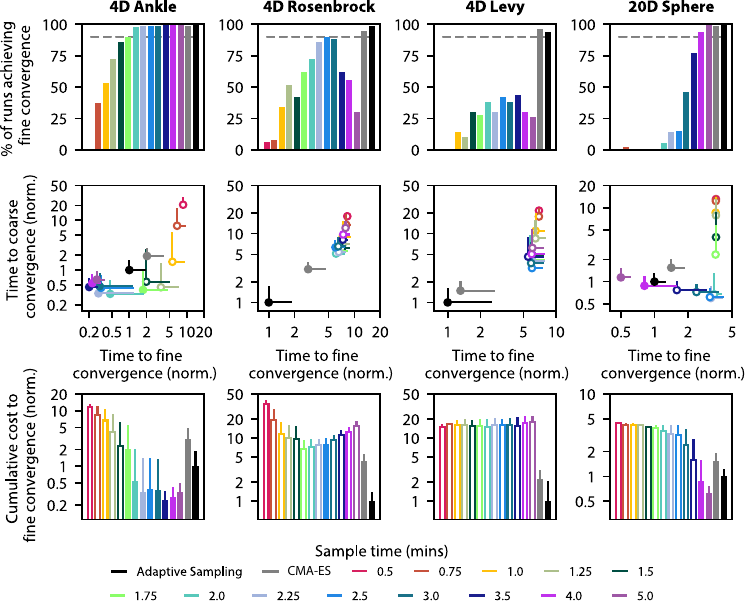}
        \caption{Performance metrics for various static sampling times Bayesian optimization (colors), the best landscape-specific CMA-ES static sampling time (gray), and AS-CMA optimization (black) runs. From left to right, the best static sampling times for CMA-ES are 4, 3, 2, and 3.5 minutes. Row 1: Convergence reliability. The gray dashed line represents the threshold fine convergence rate of 90\%; sampling strategies that did not achieve this are filled with white in the second and third rows. Row 2: Convergence time. Row 3: Convergence costs. Values in the second and third rows are normalized to AS-CMA, where smaller values indicate better performance. Whiskers for the second and third rows represent one standard deviation.}
    \label{fig:bo-metrics}
\end{figure*}

\subsection{\changeb{Comparison to KL-KG CMA-ES}}
\changeb{Static sampling KL-KG CMA-ES was similarly affected by sample time as CMA-ES: shorter measurement times resulted in fast convergence to poor-quality solutions, while longer measurement times caused slower convergence to higher-quality solutions (Figure A2). The KL-KG approach to sampling did deliver improvements over CMA-ES, although not in all landscapes. Specifically, KL-KG was clearly superior to CMA-ES in the 20D Sphere landscape, where many of its sampling strategies converged faster and more reliably than the single best static CMA-ES sampling strategy (Figure A3). In the other landscapes, the benefit of KL-KG sampling over typical CMA-ES was less apparent, delivering marginal or no benefits to convergence rate, time, and cost. Versus AS-CMA, these strengths and weaknesses of KL-KG sampling were similarly demonstrated. In the 20D sphere landscape, the best KL-KG static duration (3 mins) achieved fine convergence in 100\% of runs, accomplishing this in 13\% less time and with 25\% less cumulative cost than AS-CMA ($p<0.05$ for both comparisons). In the Ankle, Rosenbrock, and Levy landscape, the best KL-KG static duration (5, 3.5, and 3 mins) all achieved fine convergence in 91\% or more of runs, but realized this convergence in 471\%, 564\%, and 294\% more time, and with 357\%, 394\%, and 197\% more cumulative cost than AS-CMA ($p<0.05$ for all comparisons). }

\subsection{Comparison to Bayesian Optimization}
In Bayesian optimization, static sampling times produced the same trend as CMA-ES, with short measurements causing fast but poor-quality convergence and longer measurements slowing convergence while improving solution quality (Figure \ref{fig:bo_true_cost_vs_time}). Versus CMA-ES and AS-CMA, Bayesian optimization performed more favorably in simpler landscapes without parameter interaction or local minima, but worse in more complex landscapes that included either of those features (Figure \ref{fig:bo-metrics}). In favorable landscapes (Ankle and 20D Sphere) the best Bayesian optimization static sample times of 2.25 and 5 mins attained fine convergence in 99\% of all runs. Bayesian optimization achieved fine convergence in 71 and 50\% less time and with 66 and 50\% less cumulative cost than AS-CMA ($p<0.05$ for all four comparisons). In the complex landscapes (Rosenbrock and Levy) Bayesian optimization had poor fine convergence reliability, achieving the 90\% convergence reliability threshold in only one sample time (2.5 mins) in the Rosenbrock landscape and in no sample times in the Levy landscape. Because of this, all sample times were considered for best static sample time in the Levy landscape. The best static times for Rosenbrock and Levy were 2.5 mins for both landscapes, which achieved fine convergence in 90 and 42\% of runs, and in 501 and 498\% more time and with 703 and 1522\% more cumulative cost than AS-CMA ($p<0.05$ for all four comparisons). 

\subsection{\changea{Laboratory Pilot Exoskeleton Optimization Results}}
\begin{figure*}[t!]
    \centering
    \includegraphics[width=1\linewidth]{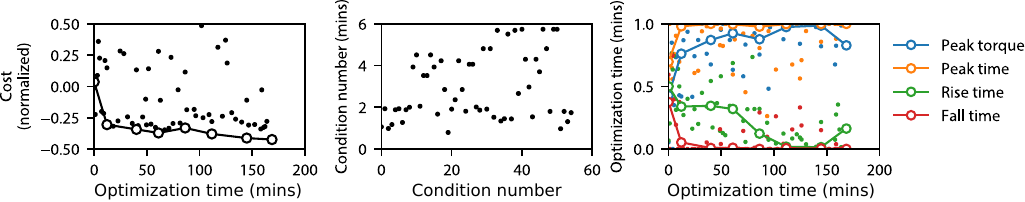}
    \caption{Results from laboratory exoskeleton assistance optimization with AS-CMA. Dots represent a single condition. White-filled circles indicate the CMA-ES mean value at each generation and are connected with a line to indicate trends in optimization.}
    \label{fig:lab-opt-results}
\end{figure*}

Results from the experimental ankle exoskeleton optimization run with AS-CMA indicate that the algorithm efficiently identified optimal assistance parameters to bring about the expected improvement in energy economy \changea{(Figure \ref{fig:lab-opt-results})}. The optimizer rapidly identified the correct direction of the minimum after one generation, evidenced by the rapid early drop in mean cost. Then, the optimizer continued to hone in on the true minimum, ending optimization with a mean energy cost reduction of 42\%, close to the expected energy cost reduction of 39\%. The optimizer correctly identified that a late peak time was most beneficial early in optimization, and later in optimization, correctly recognized that a large peak torque was also optimal. As expected, rise time did not demonstrate a strong trend, and fall time was decreased as peak time increased to ensure torque was not applied during swing. AS-CMA selected sample times covering nearly the entire allowable sample time range. 

\section{Discussion}

{\subsection{Effect of adaptive sampling \changea{in CMA-ES}}
}For efficient optimization in the presence of noise, our findings indicate it is better to vary the measurement time of each candidate instead of conducting every fitness evaluation for the same amount of time. We hypothesized that, by using simple estimates of candidate-to-candidate cost disparity, AS-CMA would identify how long each candidate should be measured, which would help to avoid under- or over-sampling candidates. If such an outcome were achieved, we expected that AS-CMA would offer more reliable, faster, and more efficient optimization. Our results support these hypotheses: across the four tested cost landscapes, AS-CMA progressively increased sampling time over optimization (Figure \ref{fig:sample_vs_exp_time}) in order to maintain a sufficiently high accuracy level (Figure \ref{fig:sorting_accuracy}), which translated into improved optimization reliability, speed, and efficiency as compared to a broad range of static sampling CMA-ES strategies (Figure \ref{fig:cma_metrics}). These results suggest that, when candidate evaluations are noisy and conducting these evaluations takes significant time, researchers should avoid static time sampling and instead should vary the measurement time over the course of optimization. \changeb{This could be accomplished with AS-CMA, which offers a straightforward sampling strategy with relatively low requirements for prior knowledge or tuning, and demonstrated promising performance across the landscapes and noise levels considered in this study.}

The favorable performance of AS-CMA in our test scenarios highlights the potential benefits of replacing static sampling with adaptive sampling in optimization tasks where CMA-ES is commonly used. As compared to static sampling CMA-ES, AS-CMA demonstrated equal convergence reliability, while achieving both coarse and fine convergence in less time and with less overall cost. Our finding that varied-time sampling is superior to static sampling aligns with prior work in noisy multiobjective optimization with evolutionary algorithms \citep{siegmund_comparative_2013,siegmund2015hybrid}, which allocated sample time proportional to the percentage of time completion of the optimization run. Our results indicating that short static sampling runs converge quickly to poor solutions, while longer static sampling runs converge to more quality solutions in more time also align with prior work in noisy optimization \citep{aizawa_scheduling_1994}. Taken together, these findings suggest that adaptive sampling could improve optimizer performance in a range of optimization tasks.

\subsection{AS-CMA versus \changea{Bayesian optimization} \changeb{and KL-KG CMA-ES} }

Compared to static sampling in Bayesian optimization, we conclude that AS-CMA is a more conservative approach, as it generalizes to a broader range of optimization scenarios but is outperformed in cost landscapes best suited to Bayesian optimization. We attribute this performance difference as being related to the complexity of the underlying cost landscape functions. The Ankle and 20D Sphere landscapes, which are unimodal and separable, were likely easier for Bayesian optimization's surrogate models to fit in comparison to the more complex Levy and Rosenbrock functions. In those complex landscapes, Bayesian optimization's surrogate model may have struggled to capture complex parameter interactions, and due to its greedy sampling strategy, may have become stuck in local minima. In general, we expect Bayesian optimization in simple landscapes (e.g. low dimensionality, few local minima, or minimal parameter interaction) will result in reliable, quick, and efficient optimization in the presence of noise, whereas AS-CMA, though equally reliable, will be slower and less efficient. Conversely, in more complex optimization landscapes, our findings suggest that AS-CMA will optimize reliability and efficiency, while Bayesian optimization may struggle to converge at all. These conclusions about the relative benefits of CMA-ES and Bayesian optimization align with other simulated human-in-the-loop optimization findings \citep{kutulakos2024simulating}, which concluded that the comparative performance of CMA-ES improves with increasing landscape complexity. Given these differences, AS-CMA may be a better choice in real-world robot optimization scenarios, where complexities such as parameter interaction, local minima, and high dimensional spaces can be common. Furthermore, the penalty for failing to converge in real-world optimization may be higher, given that reliability in these systems is often an important criterion \citep{wright_agent_2020}.

\changeb{Contrary to our expectations, KL-KG CMA-ES offered lesser performance in most cases over AS-CMA and CMA-ES, possibly due to differences in the effect of measurement noise between our study and that of \citet{groves_sequential_2018}. In that study, noise was both larger in magnitude and modeled as additive, whereas our tests used multiplicative noise of comparatively lower amplitude, chosen to represent the characteristics of human metabolic cost measurement for our exoskeleton optimization application. Furthermore, prior work using KL-KG sampling employed relatively high sampling budgets ($N_{total}=200$, $n_0=20$), but we found that such large budgets were detrimental in our cost landscapes, so we decreased them to $N_{total}=20$ and $n_0=1$. These findings suggest that the benefit of KL-KG resampling is strongly affected by the noise regime and total available sampling budget, consistent with previous observations \citep{groves_sequential_2018}. Another contributing factor to the less favorable performance of KL-KG is that it used a per-generation sampling budget of $N_{total}$, whereas the other CMA variants we tested used smaller sampling budgets of $\lambda$. In \citet{groves_sequential_2018}, comparison between KL-KG and other CMA-ES methods were made all with the same per-generation budget of $N_{total}$. At least for the noise levels seen in our study, it may be that the drawbacks for doing repeated samples of one generation of candidates $N_{total}$ times is too great relative to the value of selecting and evaluating a new generation of candidates instead. Given that KL-KG CMA-ES showed reduced effectiveness in most settings, its sensitivity to the selected per-generation sampling budget $N_{total}$, and recognizing that KL-KG sampling requires "doubling back" by sampling candidates that have already been evaluated, AS-CMA is likely a more practical choice for noisy experimental optimization problems with noise characteristics similar to what we presented here. }

\subsection{Implications for real-world optimization}
Aside from its demonstrated improvements in convergence reliability, speed, and efficiency, another important advantage of AS-CMA is that it removes a free optimization parameter, sample time, meaning that AS-CMA may be useful in optimization problems whose cost landscapes are poorly understood. Although AS-CMA outperformed CMA-ES in the four cost landscapes tested here, there may exist other landscapes where CMA-ES with an optimally-chosen static time would outperform AS-CMA. Even in this case, AS-CMA may still be the overall more time-efficient strategy because it removes the requirement of identifying the optimal static time, which would require many optimization runs with various static sample times. These duration-testing runs could also be avoided simply by guessing at a sample time, but this would risk choosing a time that is too short (worsening convergence reliability) or too long (worsening convergence time). 

Implementation of AS-CMA should be similar in complexity to CMA-ES. The approach of the original CMA-ES algorithm is unchanged; the addition of adaptive sampling requires estimates of the expected first-generation maximum and minimum costs $\hat{y}_{max}$ and $\hat{y}_{min}$ and a model of the measurement noise as a function of measurement time $\mathcal{E}(t)$. Estimating $\hat{y}_{max}$ and $\hat{y}_{min}$ is straightforward, or could even be guessed without experimentation, as imprecision in these parameters does not strongly affect optimization since they are only used for the first generation (Appendix). The noise model $\mathcal{E}(t)$ could be identified with sufficient precision by conducting several generations of CMA-ES optimization with sample time equal to the maximum desired condition time $t_{max}$. $\mathcal{E}(t)$ can then be found for each sample window size $t_i$ by estimating each condition's cost from its time-series data from $0$ to $t_i$, then identifying this estimate's error by comparing to the cost found with a window of size $t_{max}$. The errors for each time window $\epsilon_i$ can then be found by taking the standard deviation of all the errors at each time window. \changea{While correctly defining $\mathcal{E}(t)$ is more important than $\hat{y}_{max}$ and $\hat{y}_{min}$, a poorly-characterized $\mathcal{E}(t)$ is likely to still have a smaller effect on optimization outcomes than selecting a suboptimal static sampling time (Appendix).} Although AS-CMA also includes one metaparameter, the desired signal-to-noise ratio $\beta$, we find $\beta=1.3$ is a strong choice, and therefore suggest future AS-CMA users fix $\beta$ to this value to simplify optimization setup. 

\changea{AS-CMA is well-suited to noisy optimization challenges where initiating and carrying out an evaluation carries a high startup cost, whether due to human adaptation, system reconfiguration, or environmental stabilization. These startup costs make frequent switching between candidates inefficient and place a strong incentive on extracting maximum value from each sample, while also ensuring that evaluation times are short enough as to not introduce unnecessary delay. Such conditions are common in experimental domains including dexterous robotic manipulation, where task setup and feedback calibration delay each sample \citep{luo2024precise}; mobile robot navigation through unstructured terrain, where replanning and state reset impose significant overhead \citep{peng2025data}; and prosthetic or orthotic device fitting, where users require adaptation time and effort to each new setting \citep{senatore2023using}. Even in interactive machine learning systems that rely on human-in-the-loop responses, initiating new queries or training iterations can involve contextualization costs that favor smarter sample allocation \citep{mosqueira2023human}. In each of these domains—as in our motivating case of exoskeleton optimization—AS-CMA’s ability to allocate sampling time based on predicted accuracy requirement, without repeated evaluations, makes it a practical and generalizable approach for improving optimization efficiency}

\subsection{Limitations}
Situations where AS-CMA should not be used include when measurement noise is not dependent on measurement time (i.e. $\mathcal{E}(t)$ does not exist). If $\mathcal{E}(t)$ does not exist, such as if there is a strong relationship between noise and variables other than measurement time (e.g. optimization parameters), AS-CMA would not be useful. This is because AS-CMA focuses on allocating sampling time to candidates to manage the effect of noise; if candidates' measurements are unaffected by sampling time, the minimum sampling time should be used to enable rapid iteration. \changeb{Similarly, AS-CMA would not be beneficial in cases where the relationship between measurement time and measurement noise is unknown.} AS-CMA is also not recommended is contexts where CMA-ES is not well suited, such as when gradient information is available, in low-complexity cost landscapes, or in categorical optimization. 

When using AS-CMA, it is important to ensure a sufficiently accurate measurement noise-measurement time model $\mathcal{E}(t)$ is used. One type of modeling inaccuracy can occur if the experimentally-derived $\mathcal{E}(t)$ does not perfectly match the true $\mathcal{E}(t)$, however, such inaccuracy has a small effect on optimization outcomes (Appendix). The second, and more pernicious, potential issue in the formation of the noise model is model specification error. This could occur when $\mathcal{E}$ is dependent on $t$, but is also influenced by other variables, such as the ordering of candidate evaluation. To identify if there are factors other than time that strongly affect $\mathcal{E}$, we recommend characterizing $\mathcal{E}(t)$ over a range of different conditions, and checking if the error-time relationship is consistent across conditions. 

A limitation of this work is the lack of a large-scale experimental study where AS-CMA is compared to static sample time CMA-ES \changeb{with human subjects}. This was because of the high variance that is typical of stochastic optimization in the presence of noise: even though the effect size between AS-CMA and CMA-ES is large, the variance of outcome metrics such as time to convergence are also large, meaning that many trials are necessary to make statistically significant comparisons \changea{between static and adaptive sampling}. A power analysis \changeb{(Appendix)} of our simulated results from the \changea{simulated} Ankles landscape indicated that we would need $N=71$ human subject optimization runs to compare AS-CMA to the best-performing static sampling CMA-ES, which was not possible due to time and resource constraints. \changea{However, the results from the $N=100$ simulated optimization runs from each static and adaptive sampling strategy demonstrate the anticipated real-world performance of the sampling strategies in the different scenarios represented by the four cost landscapes. }

\section{Conclusion \& Future Work}
In this paper, we proposed AS-CMA, which offers a systematic approach for choosing measurement time during optimization in the presence of noise. Along six axes of performance, AS-CMA scored as well as or better than all static sampling CMA-ES strategies across four cost landscapes. Compared to Bayesian optimization with the optimal static sampling time, AS-CMA was outperformed on the simpler cost landscapes, but excelled on more complex landscapes, indicating that AS-CMA may be the more versatile option. \changeb{As compared to CMA-ES using KL-KG sampling with the optimally-chosen static sampling time, AS-CMA was slightly outperformed in one cost landscape, but demonstrated much better performance in the three other test landscapes.} These results, combined with the benefit that AS-CMA removes the requirement of selecting a measurement time, indicate that AS-CMA could be a useful tool in experimental optimization problems where noise is present. We have made this \changea{project's code}, AS-CMA demonstration code, and Python and Matlab AS-CMA packages in an ask-and-tell style interface available in an online repository for other investigators to use in future research in noisy optimization at \texttt{https://github.com/RussellMMartin/AS-CMA-ES}.

Future work should explore how measurement error can be estimated during condition evaluation, which may facilitate intelligent early stopping \changea{within each condition}, instead of the current strategy of pre-allocating each condition sample time before its evaluation. \changea{Integrating AS-CMA with a premature stopping strategy \citep{branke2016efficient} could enable earlier identification of sufficing solutions, which could further reduce optimization time.} \changeb{Future work should also investigate incorporating a known noise model into the Gaussian Process used for Bayesian optimization, which could improve surrogate fidelity and uncertainty estimation. }Additionally, given the utility of adaptive sampling for CMA-ES, it would be worthwhile to identify if a similar adaptive sampling approach could augment Bayesian optimization. \changea{Such an approach could utilize an acquisition function that incorporates expected future noise (e.g. \citet{picheny2013quantile}), while varying the sampling time (and thereby noise) as a function of optimization time, local gradient, or model uncertainty. }

\section*{Acknowledgments}
This work was supported by the National Science Foundation Graduate Research Fellowship Program (Grant number DGE-1656518). We would like to thank the Stanford Research Computing Center for providing resources that contributed to these results.

\small

\bibliographystyle{apalike}
\bibliography{as_cma_refs}

@article{aizawa_scheduling_1994,
    author = "Aizawa, Akiko N. and Wah, Benjamin W.",
    title = "Scheduling of genetic algorithms in a noisy environment",
    year = "1994",
    journal = "Evolutionary Computation",
    volume = "2",
    number = "2",
    pages = "97--122",
    doi = "10.1162/evco.1994.2.2.97",
    issn = "1063-6560, 1530-9304"
}

@article{arnold_general_2006,
    author = "Arnold, Dirk V. and Beyer, Hans-Georg",
    title = "A general noise model and its effects on evolution strategy performance",
    year = "2006",
    journal = "IEEE Transactions on Evolutionary Computation",
    volume = "10",
    number = "4",
    pages = "380--391",
    doi = "10.1109/TEVC.2005.859467",
    issn = "1941-0026"
}

@article{arnold_comparison_2003,
    author = "Arnold, Dirk V. and Beyer, Hans-Georg",
    title = "A comparison of evolution strategies with other direct search methods in the presence of noise",
    year = "2003",
    journal = "Computational Optimization and Applications",
    volume = "24",
    pages = "135--159"
}

@incollection{arnold_local_2001,
    author = "Arnold, Dirk V. and Beyer, Hans-Georg",
    editor = "Martin, Worthy N. and Spears, William M.",
    title = "Local performance of the ($\mu$/$\mu_{I}$, $\lambda$)-es in a noisy environment",
    year = "2001",
    booktitle = "Foundations of {Genetic} {Algorithms} 6",
    publisher = "Morgan Kaufmann",
    address = "San Francisco",
    pages = "127--141",
    doi = "10.1016/B978-155860734-7/50090-1",
    isbn = "978-1-55860-734-7"
}

@incollection{bergener_parameter_2001,
    author = "Bergener, Thomas and Bruckhoff, Carsten and Igel, Christian",
    title = "Parameter optimization for visual obstacle detection using a derandomized evolution strategy",
    year = "2001",
    booktitle = "Imaging and {Vision} {Systems}: {Theory}, {Assessment} and {Applications}, volume 9 of {Advances} in {Computation}: {Theory} and {Practice}",
    publisher = "NOVA Science Books",
    address = "Huntington, NY, USA",
    pages = "265--279"
}

@inproceedings{branke_efficient_2001,
    author = {Branke, J{\"u}rgen and Schmidt, Christian and Schmeck, Hartmut},
    title = "Efficient fitness estimation in noisy environments",
    year = "2001",
    booktitle = "Proceedings of the Genetic and Evolutionary Computation Conference (GECCO)",
    publisher = "Morgan Kaufmann Publishers Inc.",
    address = "San Francisco, CA, USA",
    pages = "243--250",
    isbn = "978-1-55860-774-3"
}

@inproceedings{branke2003selection,
    author = {Branke, J{\"u}rgen and Schmidt, Christian},
    title = "Selection in the presence of noise",
    booktitle = "Proceedings of the Genetic and Evolutionary Computation Conference (GECCO)",
    pages = "766--777",
    year = "2003",
    organization = "Springer"
}

@inproceedings{di_pietro_applying_2004,
    author = "Di Pietro, A. and While, L. and Barone, L.",
    title = "Applying evolutionary algorithms to problems with noisy, time-consuming fitness functions",
    year = "2004",
    booktitle = "IEEE Congress on Evolutionary Computation (CEC)", 
    pages = "1254--1261",
    doi = "10.1109/CEC.2004.1331041"
}

@article{ding_human---loop_2018,
    author = "Ding, Ye and Kim, Myunghee and Kuindersma, Scott and Walsh, Conor J.",
    title = "Human-in-the-loop optimization of hip assistance with a soft exosuit during walking",
    year = "2018",
    journal = "Science Robotics",
    volume = "3",
    number = "15",
    pages = "eaar5438",
    doi = "10.1126/scirobotics.aar5438",
    issn = "2470-9476"
}

@article{ding_effect_2016,
    author = "Ding, Ye and Panizzolo, Fausto A. and Siviy, Christopher and Malcolm, Philippe and Galiana, Ignacio and Holt, Kenneth G. and Walsh, Conor J.",
    title = "Effect of timing of hip extension assistance during loaded walking with a soft exosuit",
    year = "2016",
    journal = "Journal of NeuroEngineering and Rehabilitation",
    volume = "13",
    number = "1",
    pages = "87",
    doi = "10.1186/s12984-016-0196-8",
    issn = "1743-0003"
}

@article{dixon1978global,
    author = "Dixon, Laurence Charles Ward",
    title = "The global optimization problem: an introduction",
    journal = "Towards Global Optimiation 2",
    pages = "1--15",
    year = "1978",
    publisher = "North-Holland"
}

@article{franks_comparing_2021,
    author = "Franks, Patrick W. and Bryan, Gwendolyn M. and Martin, Russell M. and Reyes, Ricardo and Lakmazaheri, Ava C. and Collins, Steven H.",
    title = "Comparing optimized exoskeleton assistance of the hip, knee, and ankle in single and multi-joint configurations",
    year = "2021",
    journal = "Wearable Technologies",
    volume = "2",
    doi = "10.1017/wtc.2021.14",
    issn = "2631-7176"
}

@article{grimmer_comparison_2019,
    author = "Grimmer, Martin and Quinlivan, Brendan T. and Lee, Sangjun and Malcolm, Philippe and Rossi, Denise Martineli and Siviy, Christopher and Walsh, Conor J.",
    title = "Comparison of the human-exosuit interaction using ankle moment and ankle positive power inspired walking assistance",
    year = "2019",
    journal = "Journal of Biomechanics",
    volume = "83",
    pages = "76--84",
    doi = "10.1016/j.jbiomech.2018.11.023",
    issn = "00219290"
}

@inproceedings{groves_sequential_2018,
    author = "Groves, Matthew and Branke, Juergen",
    title = "Sequential sampling for noisy optimisation with {CMA}-{ES}",
    year = "2018",
    booktitle = "Proceedings of the Genetic and Evolutionary Computation Conference (GECCO)",
    publisher = "Association for Computing Machinery",
    address = "New York, NY, USA",
    pages = "1023--1030",
    doi = "10.1145/3205455.3205559",
    isbn = "978-1-4503-5618-3"
}

@article{handford_energy-optimal_2018,
    author = "Handford, Matthew L. and Srinivasan, Manoj",
    title = "Energy-optimal human walking with feedback-controlled robotic prostheses: a computational study",
    year = "2018",
    journal = "IEEE Transactions on Neural Systems and Rehabilitation Engineering",
    volume = "26",
    number = "9",
    pages = "1773--1782",
    doi = "10.1109/TNSRE.2018.2858204",
    issn = "1558-0210"
}

@article{hansen_reducing_2003,
    title={Reducing the time complexity of the derandomized evolution strategy with covariance matrix adaptation ({CMA}-{ES})},
    author={Hansen, Nikolaus and M{\"u}ller, Sibylle D and Koumoutsakos, Petros},
    year = "2003",
    journal = "Evolutionary Computation",
    volume = "11",
    number = "1",
    pages = "1--18",
    doi = "10.1162/106365603321828970",
    issn = "1063-6560, 1530-9304"
}

@article{hansen_method_2008,
    author = "Hansen, Nikolaus and Niederberger, Andr?? S. P. and Guzzella, Lino and Koumoutsakos, Petros",
    title = "A method for handling uncertainty in evolutionary optimization with an application to feedback control of combustion",
    year = "2008",
    journal = "IEEE Transactions on Evolutionary Computation",
    volume = "13",
    number = "1",
    pages = "180--197",
    doi = "10.1109/TEVC.2008.924423",
    issn = "1941-0026"
}

@article{hansen2016cma,
    author = "Hansen, Nikolaus",
    title = "The {CMA} evolution strategy: a tutorial",
    journal = "arXiv preprint arXiv:1604.00772",
    year = "2016"
}

@inproceedings{hill_online_2020,
    author = "Hill, Ashley and Laneurit, Jean and Lenain, Roland and Lucet, Eric",
    title = "Online gain setting method for path tracking using {CMA}-{ES}: application to off-road mobile robot control",
    year = "2020",
    booktitle = "IEEE/RSJ International Conference on Intelligent Robots and Systems (IROS)",
    pages = "7697--7702",
    doi = "10.1109/IROS45743.2020.9340830"
}

@article{hu_evolution_2019,
    author = "Hu, Yingbai and Wu, Xinyu and Geng, Peng and Li, Zhijun",
    title = "Evolution strategies learning with variable impedance control for grasping under uncertainty",
    year = "2019",
    journal = "IEEE Transactions on Industrial Electronics",
    volume = "66",
    number = "10",
    pages = "7788--7799",
    doi = "10.1109/TIE.2018.2884240",
    issn = "1557-9948"
}

@article{kutulakos2024simulating,
    author = "Kutulakos, Zoe B and Slade, Patrick",
    title = "Simulating human-in-the-loop optimization of exoskeleton assistance to compare optimization algorithm performance",
    journal = "bioRxiv",
    pages = "2024--04",
    year = "2024",
    publisher = "Cold Spring Harbor Laboratory"
}

@article{laguna_experimental_2005,
    author = "Laguna, Manuel and Marti, Rafael",
    title = "Experimental testing of advanced scatter search designs for global optimization of multimodal functions",
    year = "2005",
    journal = "Journal of Global Optimization ",
    volume = "33",
    pages = "235--255"
}

@article{lakmazaheri2024optimizing,
    author = "Lakmazaheri, Ava C and Song, Seungmoon and Vuong, Brian B and Biskner, Blake and Kado, Deborah M and Collins, Steven H",
    title = "Optimizing exoskeleton assistance to improve walking speed and energy economy for older adults",
    journal = "Journal of NeuroEngineering and Rehabilitation",
    volume = "21",
    number = "1",
    pages = "1",
    year = "2024",
    publisher = "Springer"
}

@article{malcolm_simple_2013,
    author = "Malcolm, Philippe and Derave, Wim and Galle, Samuel and Clercq, Dirk De",
    title = "A simple exoskeleton that assists plantarflexion can reduce the metabolic cost of human walking",
    year = "2013",
    journal = "PLoS One",
    volume = "8",
    number = "2",
    pages = "e56137",
    doi = "10.1371/journal.pone.0056137",
    issn = "1932-6203"
}

@article{nomura_cmaes_2024,
    author = "Nomura, Masahiro and Shibata, Masashi",
    title = "cmaes: a simple yet practical python library for {CMA}-{ES}",
    year = "2024",
    journal = "arXiv"
}

@article{poggensee_how_2021,
    author = "Poggensee, Katherine L. and Collins, Steven H.",
    title = "How adaptation, training, and customization contribute to benefits from exoskeleton assistance",
    year = "2021",
    journal = "Science Robotics",
    volume = "6",
    number = "58",
    pages = "eabf1078",
    doi = "10.1126/scirobotics.abf1078"
}

@article{rakshit_noisy_2017,
    author = "Rakshit, Pratyusha and Konar, Amit and Das, Swagatam",
    title = "Noisy evolutionary optimization algorithms: a comprehensive survey",
    year = "2017",
    journal = "Swarm and Evolutionary Computation",
    volume = "33",
    pages = "18--45",
    doi = "10.1016/j.swevo.2016.09.002",
    issn = "2210-6502"
}

@inproceedings{sano_optimization_2002,
    author = "Sano, Y. and Kita, H.",
    title = "Optimization of noisy fitness functions by means of genetic algorithms using history of search with test of estimation",
    year = "2002",
    booktitle = "IEEE Congress on Evolutionary Computation (CEC)",
    volume = "1",
    pages = "360--365 vol.1",
    doi = "10.1109/CEC.2002.1006261"
}

@article{shafii_learning_2015,
    author = "Shafii, Nima and Lau, Nuno and Reis, Luis Paulo",
    title = "Learning to walk fast: optimized hip height movement for simulated and real humanoid robots",
    year = "2015",
    journal = "Journal of Intelligent \& Robotic Systems",
    volume = "80",
    number = "3-4",
    pages = "555--571",
    doi = "10.1007/s10846-015-0191-5",
    issn = "0921-0296, 1573-0409"
}

@article{sharifzadeh_maneuverable_2021,
    author = "Sharifzadeh, Mohammad and Jiang, Yuhao and Lafmejani, Amir Salimi and Nichols, Kevin and Aukes, Daniel",
    title = "Maneuverable gait selection for a novel fish-inspired robot using a {CMA}-{ES}-assisted workflow",
    year = "2021",
    journal = "Bioinspiration \& Biomimetics",
    volume = "16",
    number = "5",
    pages = "056017",
    doi = "10.1088/1748-3190/ac165d",
    issn = "1748-3182, 1748-3190"
}

@inproceedings{siegmund2015hybrid,
    author = "Siegmund, Florian and Ng, Amos HC and Deb, Kalyanmoy",
    title = "Hybrid dynamic resampling for guided evolutionary multi-objective optimization",
    booktitle = "International Conference on Evolutionary Multi-Criterion Optimization",
    pages = "366--380",
    year = "2015",
    organization = "Springer"
}

@inproceedings{siegmund_comparative_2013,
    author = "Siegmund, Florian and Ng, Amos H.C. and Deb, Kalyanmoy",
    title = "A comparative study of dynamic resampling strategies for guided evolutionary multi-objective optimization",
    year = "2013",
    booktitle = "IEEE Congress on Evolutionary Computation (CEC)",
    pages = "1826--1835",
    doi = "10.1109/CEC.2013.6557782"
}

@article{slade_human---loop_2024,
    author = "Slade, Patrick and Atkeson, Christopher and Donelan, J. Maxwell and Houdijk, Han and Ingraham, Kimberly A. and Kim, Myunghee and Kong, Kyoungchul and Poggensee, Katherine L. and Riener, Robert and Steinert, Martin and Zhang, Juanjuan and Collins, Steven H.",
    title = "On human-in-the-loop optimization of human-robot interaction",
    year = "2024",
    journal = "Nature",
    volume = "633",
    number = "8031",
    pages = "779--788",
    doi = "10.1038/s41586-024-07697-2",
    issn = "0028-0836, 1476-4687"
}

@article{slade_personalizing_2022,
    author = "Slade, Patrick and Kochenderfer, Mykel J. and Delp, Scott L. and Collins, Steven H.",
    title = "Personalizing exoskeleton assistance while walking in the real world",
    year = "2022",
    journal = "Nature",
    volume = "610",
    number = "7931",
    pages = "277--282",
    doi = "10.1038/s41586-022-05191-1",
    issn = "1476-4687",
    copyright = "2022 The Author(s)"
}

@article{wright_agent_2020,
    author = "Wright, Julia L. and Chen, Jessie Y. C. and Lakhmani, Shan G.",
    title = "Agent transparency and reliability in human-robot interaction: the influence on user confidence and perceived reliability",
    year = "2020",
    journal = "IEEE Transactions on Human-Machine Systems",
    volume = "50",
    number = "3",
    pages = "254--263",
    doi = "10.1109/THMS.2019.2925717",
    issn = "2168-2305"
}

@article{zhang2017human,
    author = "Zhang, Juanjuan and Fiers, Pieter and Witte, Kirby A and Jackson, Rachel W and Poggensee, Katherine L and Atkeson, Christopher G and Collins, Steven H",
    title = "Human-in-the-loop optimization of exoskeleton assistance during walking",
    journal = "Science",
    volume = "356",
    number = "6344",
    pages = "1280--1284",
    year = "2017",
    publisher = "American Association for the Advancement of Science"
}

@book{kochenderfer2019algorithms,
    author = "Kochenderfer, MJ",
    title = "Algorithms for optimization",
    year = "2019",
    publisher = "The MIT Press Cambridge"
}

@article{rosenbrock1960automatic,
    author = "Rosenbrock, HoHo",
    title = "An automatic method for finding the greatest or least value of a function",
    journal = "The Computer Journal ",
    volume = "3",
    number = "3",
    pages = "175--184",
    year = "1960",
    publisher = "Oxford University Press"
}

@Misc{nogueira2014bayesian,
    author = "Nogueira, Fernando",
    title = "Bayesian optimization: open source constrained global optimization tool for python",
    year = "2014"
}

@article{spearman1961proof,
    author = "Spearman, Charles",
    title = "The proof and measurement of association between two things.",
    year = "1904",
    journal = "American Journal of Psychology",
    pages = "72--101",
    volume = "15",
    number = "1"
}

@article{wang2020sample,
  title={Sample size estimation in clinical research: from randomized controlled trials to observational studies},
  author={Wang, Xiaofeng and Ji, Xinge},
  journal={Chest},
  volume={158},
  number={1},
  pages={S12--S20},
  year={2020},
  publisher={Elsevier}
}

@article{zhu2020ingredients,
  title={The ingredients of real-world robotic reinforcement learning},
  author={Zhu, Henry and Yu, Justin and Gupta, Abhishek and Shah, Dhruv and Hartikainen, Kristian and Singh, Avi and Kumar, Vikash and Levine, Sergey},
  journal={arXiv preprint arXiv:2004.12570},
  year={2020}
}

@inproceedings{kang2019generalization,
  title={Generalization through simulation: Integrating simulated and real data into deep reinforcement learning for vision-based autonomous flight},
  author={Kang, Katie and Belkhale, Suneel and Kahn, Gregory and Abbeel, Pieter and Levine, Sergey},
  booktitle={IEEE International Conference on Robotics and Automation (ICRA)},
  pages={6008--6014},
  year={2019},
}

@article{zhang2019vr,
  title={Vr-goggles for robots: real-to-sim domain adaptation for visual control},
  author={Zhang, Jingwei and Tai, Lei and Yun, Peng and Xiong, Yufeng and Liu, Ming and Boedecker, Joschka and Burgard, Wolfram},
  journal={IEEE Robotics and Automation Letters},
  volume={4},
  number={2},
  pages={1148--1155},
  year={2019},
  publisher={IEEE}
}

@inproceedings{nishida2018psa,
  title={Psa-cma-es: Cma-es with population size adaptation},
  author={Nishida, Kouhei and Akimoto, Youhei},
  booktitle={Proceedings of the Genetic and Evolutionary Computation Conference (GECCO)},
  pages={865--872},
  year={2018}
}

@article{selinger2014estimating,
  title={Estimating instantaneous energetic cost during non-steady-state gait},
  author={Selinger, Jessica C and Donelan, J Maxwell},
  journal={Journal of Applied Physiology},
  volume={117},
  number={11},
  pages={1406--1415},
  year={2014},
}

@article{picheny2013quantile,
  title={Quantile-based optimization of noisy computer experiments with tunable precision},
  author={Picheny, Victor and Ginsbourger, David and Richet, Yann and Caplin, Gregory},
  journal={Technometrics},
  volume={55},
  number={1},
  pages={2--13},
  year={2013},
}

@article{srinivas2009gaussian,
  title={Gaussian process optimization in the bandit setting: No regret and experimental design},
  author={Srinivas, Niranjan and Krause, Andreas and Kakade, Sham M and Seeger, Matthias},
  journal={arXiv preprint arXiv:0912.3995},
  year={2009}
}

@book{forrester2008engineering,
  title={Engineering design via surrogate modelling: a practical guide},
  author={Forrester, Alexander and Sobester, Andras and Keane, Andy},
  year={2008},
  publisher={John Wiley \& Sons}
}

@article{xiao2014simulation,
  title={Simulation optimization using genetic algorithms with optimal computing budget allocation},
  author={Xiao, Hui and Lee, Loo Hay},
  journal={Simulation},
  volume={90},
  number={10},
  pages={1146--1157},
  year={2014},
  publisher={Sage Publications Sage UK: London, England}
}

@article{branke2016efficient,
  title={Efficient use of partially converged simulations in evolutionary optimization},
  author={Branke, Juergen and Asafuddoula, Md and Bhattacharjee, Kalyan Shankar and Ray, Tapabrata},
  journal={IEEE Transactions on Evolutionary Computation},
  volume={21},
  number={1},
  pages={52--64},
  year={2016},
  publisher={IEEE}
}

@Inbook{Bartz-Beielstein2007,
author="Bartz-Beielstein, Thomas
and Blum, Daniel
and Branke, J{\"u}rgen",
title="Particle Swarm Optimization and Sequential Sampling in Noisy Environments",
bookTitle="Metaheuristics: Progress in Complex Systems Optimization",
year="2007",
publisher="Springer US",
address="Boston, MA",
pages="261--273",
isbn="978-0-387-71921-4",
doi="10.1007/978-0-387-71921-4_14",
url="https://doi.org/10.1007/978-0-387-71921-4_14"
}

@inproceedings{schutte2024improving,
  title={Improving metaheuristic efficiency for stochastic optimization by sequential predictive sampling},
  author={Schutte, Noah and Postek, Krzysztof and Yorke-Smith, Neil},
  booktitle={CPAIOR},
  pages={158--175},
  year={2024},
  organization={Springer}
}

@inproceedings{heidrich2009hoeffding,
  title={Hoeffding and Bernstein races for selecting policies in evolutionary direct policy search},
  author={Heidrich-Meisner, Verena and Igel, Christian},
  booktitle={Proceedings of the 26th Annual International Conference on Machine Learning},
  pages={401--408},
  year={2009}
}

@inproceedings{schmidt2006integrating,
  title={Integrating techniques from statistical ranking into evolutionary algorithms},
  author={Schmidt, Christian and Branke, J{\"u}rgen and Chick, Stephen E},
  booktitle={Applications of Evolutionary Computing, Budapest, Hungary, April 10-12, 2006. Proceedings},
  pages={752--763},
  year={2006},
  organization={Springer}
}

@article{luo2024precise,
  title={Precise and Dexterous Robotic Manipulation via Human-in-the-Loop Reinforcement Learning},
  author={Luo, Jianlan and Xu, Charles and Wu, Jeffrey and Levine, Sergey},
  journal={arXiv preprint arXiv:2410.21845},
year={2024}
}

@article{peng2025data,
  title={Data-Efficient Learning from Human Interventions for Mobile Robots},
  author={Peng, Zhenghao and Liu, Zhizheng and Zhou, Bolei},
  journal={arXiv preprint arXiv:2503.04969},
year={2025}
}

@article{senatore2023using,
  title={Using human-in-the-loop optimization for guiding manual prosthesis adjustments: a proof-of-concept study},
  author={Senatore, Siena C and Takahashi, Kota Z and Malcolm, Philippe},
  journal={Frontiers in Robotics and AI},
  volume={10},
  pages={1183170},
  year={2023},
  publisher={Frontiers Media SA}
}

@article{mosqueira2023human,
  title={Human-in-the-loop machine learning: a state of the art},
  author={Mosqueira-Rey, Eduardo and Hern{\'a}ndez-Pereira, Elena and Alonso-R{\'\i}os, David and Bobes-Bascar{\'a}n, Jos{\'e} and Fern{\'a}ndez-Leal, {\'A}ngel},
  journal={Artificial Intelligence Review},
  volume={56},
  number={4},
  pages={3005--3054},
  year={2023},
  publisher={Springer}
}

\newpage 

\section*{Appendix}
\textit{Improving CMA-ES convergence speed, efficiency, and reliability in noisy robot optimization problems}
\textit{Martin and Collins, 2026}
\pagestyle{plain} \pagenumbering{gobble} 

\subsection*{Importance of accurate $\hat{y}$ and $\mathcal{E}(t)$}
To test the importance of accurate estimates of AS-CMA landscape-related parameters ($\hat{y}_{max}$, $\hat{y}_{min}$, and $\mathcal{E}(t)$), we conducted 5 tests on the Ankle landscape, introducing a different amount of error on each test. Error was applied as $parameter \leftarrow parameter \times (1 + error)$, where $error$ for that test was one of $[0, \pm0.1, \pm0.25]$, and $parameter$ was either $[\hat{y}_{max}, \hat{y}_{min}]$ or $\mathcal{E}(t)$. Each test had $N=100$ runs. Results (Figure A1) showed that all levels of error introduced to $\hat{y}_{max}$ and $\hat{y}_{min}$ had a negligible effect on the trajectory of the optimizer's mean until very fine levels of convergence. For error in $\mathcal{E}(t)$,  under-estimations of $\mathcal{E}(t)$ moderately affected fine convergence, while over-estimations moderately affect coarse convergence. However, the overall effect of misestimating either $\hat{y}$ or $\mathcal{E}(t)$ was small compared to the effect of picking a suboptimal static sampling time. 
\begin{figure}[h!]
    \centering
    \includegraphics[width=1\linewidth]{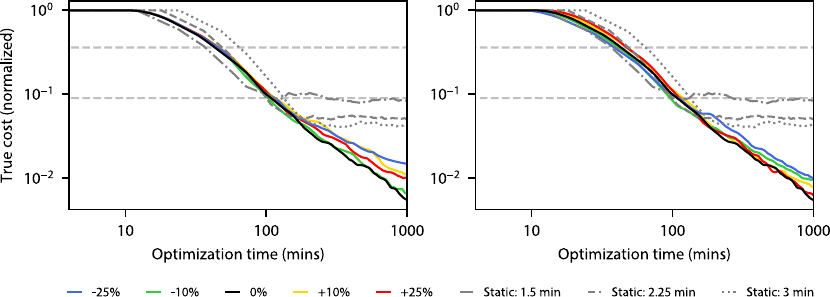}
    \captionsetup{labelformat=empty}
    \caption*{Figure A1: Results from introducing small and large errors to $\hat{y}_{max}$ and $\hat{y}_{min}$ (left), or to $\mathcal{E}(t)$ (right). Percentages in the legend indicate the magnitude of error introduced. Three static sample times CMA-ES results (gray) are also shown for reference; 2.25 min sample time was identified as the best static sample time for this cost landscape. }
\end{figure}

\newpage
\subsection*{\changeb{Comparison of AS-CMA to KL-KG CMA-ES}}
\begin{figure}[h!]
    \centering
    \includegraphics[width=1\linewidth]{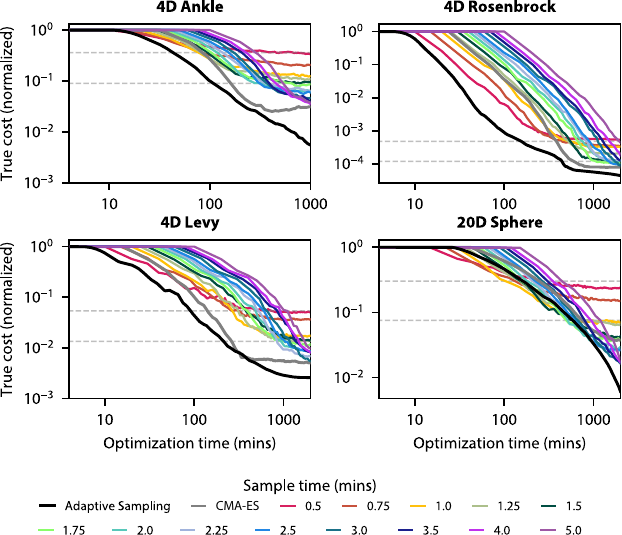}
    \captionsetup{labelformat=empty}
    \caption*{\changeb{Figure A2: True cost of estimated optimum over KL-KG CMA-ES comparison experiment (each line represents average of $N=100$ runs). The best CMA-ES static sample time (gray line) was 4, 3, 2, and 3.5 minutes for Ankle, Rosenbrock, Levy, and 20D Sphere, respectively. The higher and lower dashed lines indicate the coarse and fine convergence thresholds.  The vertical axes are normalized to the cost of the AS-CMA and CMA-ES optimization starting points, which were the same across sampling strategies}.}
\end{figure}

\begin{figure}[h!]
    \centering
    \includegraphics[width=1\linewidth]{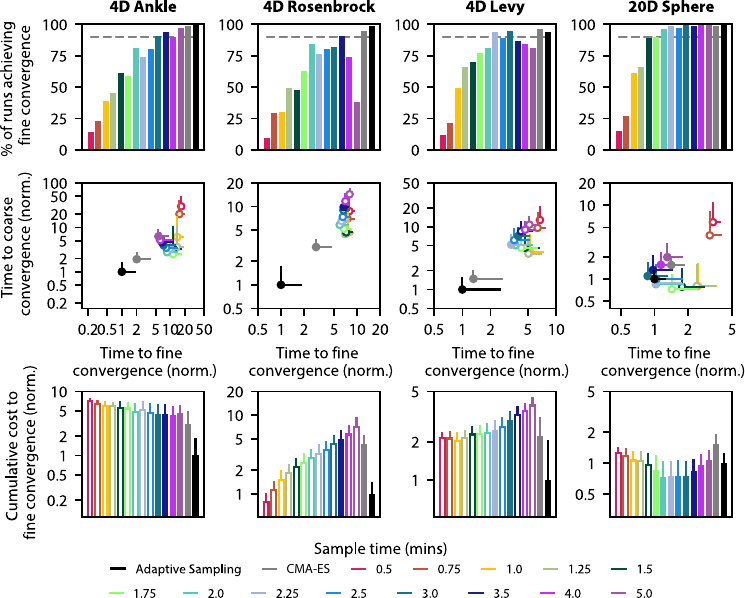}
    \captionsetup{labelformat=empty}
    \caption*{\changeb{Figure A3: Performance metrics for various static sampling times using KL-KG CMA-ES (colors), the best landscape-specific CMA-ES static sampling time (gray), and AS-CMA optimization (black) runs. From left to right, the best static sampling times for CMA-ES are 4, 3, 2, and 3.5 minutes. Row 1: Convergence reliability. The gray dashed line represents the threshold fine convergence rate of 90\%; sampling strategies that did not achieve this are filled with white in the second and third rows. Row 2: Convergence time. Row 3: Convergence costs. Values in the second and third rows are normalized to AS-CMA, where smaller values indicate better performance. Whiskers for the second and third rows represent one standard deviation. }}
\end{figure}

\clearpage

\subsection*{\changea{Percent change from Bayesian Optimization/CMA-ES to AS-CMA}}

\changea{Tables A1 and A2 give quantiative comparisons between CMA-ES/Bayesian Optimization and AS-CMA. The best CMA-ES static sample time for Ankle, Rosenbrock, Levy, and 20D Sphere are 4, 3, 2, and 3.5 minutes, respectively. The best static times for Bayesian optimization are 2.25, 2.5, 2.5, and 5 minutes.
}
\captionsetup{type=table}
\caption*{\textbf{\changea{Table A1:}} \changea{Percentage change in convergence rate, time, and cost  across fine and coarse thresholds when changing from CMA-ES to AS-CMA. All values represent percent change in performance of AS-CMA relative to the best static sampling time CMA-ES. Improved performance by AS-CMA is indicated by a positive value in the Convergence Rate columns and a negative value in the Convergence Time and Cost columns.} }
\begin{tabular}{@{}lcccccc@{}}
\toprule
\textbf{Landscape} & \multicolumn{2}{c}{\textbf{Conv. Rate}} & \multicolumn{2}{c}{\textbf{Conv. Time}} & \multicolumn{2}{c}{\textbf{Conv. Cost}} \\
& Fine & Coarse & Fine & Coarse & Fine & Coarse \\
\midrule
4D Ankle        & 1\%  & 0\%  & -51\% & -48\% & -51\% & -50\% \\
4D Rosenbrock   & 4\%  & 0\%  & -65\% & -67\% & -76\% & -76\% \\
4D Levy         & -2\% & 0\%  & -24\% & -32\% & -37\% & -46\% \\
20D Sphere      & 1\%  & 0\%  & -29\% & -35\% & -29\% & -36\% \\
\bottomrule
\end{tabular}
\vspace{1ex}

\captionsetup{type=table}
\caption*{\textbf{\changea{Table A2:}} \changea{Percentage change in convergence rate, time, and cost  across fine and coarse thresholds when changing from Bayesian optimization to AS-CMA. All values represent percent change in performance of AS-CMA relative to the best static sampling time Bayesian optimization. Improved performance by AS-CMA is indicated by a positive value in the Convergence Rate columns and a negative value in the Convergence Time and Cost columns.} }

\begin{tabular}{@{}lcccccc@{}}
\toprule
\textbf{Landscape} & \multicolumn{2}{c}{\textbf{Rate}} & \multicolumn{2}{c}{\textbf{Time}} & \multicolumn{2}{c}{\textbf{Cost}} \\
& Fine & Coarse & Fine & Coarse & Fine & Coarse \\
\midrule
4D Ankle        & 1\%   & 0\%   & 245\%  & 194\%  & 190\%  & 162\% \\
4D Rosenbrock   & 10\%  & 0\%   & -83\%  & -84\%  & -88\%  & -87\% \\
4D Levy         & 124\% & 0\%   & -83\%  & -68\%  & -94\%  & -83\% \\
20D Sphere      & 0\%   & 0\%   & 61\%   & -16\%  & 24\%   & -24\% \\
\bottomrule
\end{tabular}

\captionsetup{type=table}
\caption*{\textbf{\changeb{Table A3:}} \changeb{Percentage change in convergence rate, time, and cost  across fine and coarse thresholds when changing from KL-KG CMA-ES to AS-CMA. All values represent percent change in performance of AS-CMA relative to the best static sampling time KL-KG CMA-ES optimization. Improved performance by AS-CMA is indicated by a positive value in the Convergence Rate columns and a negative value in the Convergence Time and Cost columns.} }

\begin{tabular}{@{}lcccccc@{}}
\toprule
\textbf{Landscape} & \multicolumn{2}{c}{\textbf{Rate}} & \multicolumn{2}{c}{\textbf{Time}} & \multicolumn{2}{c}{\textbf{Cost}} \\
& Fine & Coarse & Fine & Coarse & Fine & Coarse \\
\midrule
4D Ankle        & 3\%   & 0\%   & -82\%  & -84\%  & -78\%  & -83\% \\
4D Rosenbrock   & 9\%   & 2\%   & -85\%  & -90\%  & -80\%  & -80\% \\
4D Levy         & 1\%   & -1\%  & -75\%  & -86\%  & -66\%  & -79\% \\
20D Sphere      & 0\%   & 0\%   & 15\%   & -9\%  & 33 \%   &  4\% \\
\bottomrule
\end{tabular}

\clearpage
\section*{Power analysis for potential large-scale human subject experiments}
\changeb{Our simulated ankle exoskeleton optimization indicate that the best static sample time CMA-ES (2.25 minutes) and AS-CMA require $233 \pm 226$ and $115 \pm 107$ minutes (mean $\pm$ standard deviation) of optimization to achieve fine convergence, respectively. With a significance level of $\alpha = 0.05$ and 80\% power ($\beta = 0.2$), we would require a sample size of at least $N=71$ participants to detect this difference in a gait study \citep{wang2020sample}. 
}\end{document}